\documentclass[10pt, letter, onecolumn]{arxiv}

\usepackage{kantlipsum, lipsum}
\usepackage{dm-colors}
\usepackage{amsmath}
\usepackage{pstricks, pst-node}
\usepackage{verbatim}
\usepackage{multirow}
\usepackage{scalerel}
\usepackage{booktabs}
\usepackage{enumitem}
\usepackage{xspace}
\usepackage{bm}
\usepackage{bbm}
\usepackage{mathtools}
\usepackage{soul}
\usepackage{epsfig}
\usepackage{graphicx}
\usepackage{csquotes}
\usepackage{setspace}
\usepackage{colortbl}
\usepackage{threeparttable}
\usepackage{tabularx,ragged2e}
\usepackage{placeins}
\usepackage[hang,flushmargin,symbol]{footmisc}
\usepackage{longtable}
\usepackage{multirow}
\usepackage{makecell}
\usepackage{tabularx}
\usepackage{float}
\newcolumntype{C}[1]{>{\centering\arraybackslash}m{#1}}

\usepackage{varioref}
\usepackage[pagebackref=false,breaklinks=false,%
            colorlinks=true,bookmarks=true,citecolor=ourdarkblue,%
            urlcolor=ourdarkblue,linkcolor=ourdarkblue]{hyperref}
\usepackage[noabbrev,capitalize]{cleveref}
\usepackage{etoc}
\usepackage{lineno}

\usepackage{subcaption}
\usepackage{caption}
\usepackage{threeparttable}

\graphicspath{{figures/}}

\usepackage[misc]{ifsym}

\title{\Large{Self-supervised Anomaly Detection Pretraining \\Enhances Long-tail ECG Diagnosis}}

\vspace{1cm}
\author[$\ast$,1,2]{Aofan Jiang} 
\author[$\ast$,1,2]{Chaoqin Huang}
\author[$\ast$,3]{Qing Cao}
\author[3]{Yuchen Xu}
\author[3]{Zi Zeng}
\author[3]{Kang Chen}
\author[1,2, \Letter]{\\ \vspace{0.1cm} Ya Zhang}
\author[1,2, \Letter]{Yanfeng Wang}

\affil[1]{\normalsize Shanghai Jiao Tong University \hspace{1cm}}

\affil[2]{\normalsize Shanghai AI Laboratory  \hspace{1cm}\ \ \ \ \ \ \ \ \ \ \ \ \ \ \ \ \ \ \ \ \ \ \ \ \ \ \ \ \ \ \ \ \ \ \ \ }

\affil[3]{\normalsize Ruijin Hospital, Shanghai Jiao Tong University School of Medicine}

\renewcommand{\correspondingauthor}[1]{$\ast$~Equal contributions.}

\begin{document}

\begin{abstract}
Current computer-aided ECG diagnostic systems struggle with the underdetection of rare but critical cardiac anomalies due to the imbalanced nature of ECG datasets. This study introduces a novel approach using self-supervised anomaly detection pretraining to address this limitation. The anomaly detection model is specifically designed to detect and localize subtle deviations from normal cardiac patterns, capturing the nuanced details essential for accurate ECG interpretation. Validated on an extensive dataset of over one million ECG records from clinical practice, characterized by a long-tail distribution across 116 distinct categories, the anomaly detection-pretrained ECG diagnostic model has demonstrated a significant improvement in overall accuracy. Notably, our approach yielded a 94.7\% AUROC, 92.2\% sensitivity, and 92.5\% specificity for rare ECG types, significantly outperforming traditional methods and narrowing the performance gap with common ECG types. The integration of anomaly detection pretraining into ECG analysis represents a substantial contribution to the field, addressing the long-standing challenge of long-tail data distributions in clinical diagnostics. Furthermore, prospective validation in real-world clinical settings revealed that our AI-driven approach enhances diagnostic efficiency, precision, and completeness by 32\%, 6.7\%, and 11.8\% respectively, when compared to standard practices. This advancement marks a pivotal step forward in the integration of AI within clinical cardiology, with particularly profound implications for emergency care, where rapid and accurate ECG interpretation is crucial. The contributions of this study not only push the boundaries of current ECG diagnostic capabilities but also lay the groundwork for more reliable and accessible cardiovascular care.
\end{abstract}

\maketitle

\section{Introduction}

In the ever-evolving field of healthcare, the demand for accurate and timely diagnosis of cardiac conditions through non-invasive methods has become increasingly urgent~\cite{wearble, ecgsurvey}. Among these methods, the electrocardiogram (ECG) remains a pivotal tool, offering critical insights into the heart’s electrical activity. Despite its widespread use and cost-effectiveness, the diagnostic accuracy of ECG interpretation remains suboptimal, even among well-trained medical professionals.  A recent meta-analysis of 78 studies~\cite{10.1001/jamainternmed.2020.3989} reveals that diagnostic accuracy rates are only 42.0\%, 55.8\%, and 68.5\%  for medical students,  residents, and non-cardiovascular specialist physicians, respectively. These figures are likely lower in regions with fewer resources, highlighting an urgent need for innovative solutions to enhance diagnostic precision and accessibility in cardiovascular care.

\begin{figure*}[p]
    \centering
    \begin{subfigure}[t]{\textwidth}
    \centering
    \caption{a. Long-tail distribution of cardiac types across the entire dataset.}
    \includegraphics[width=\linewidth]{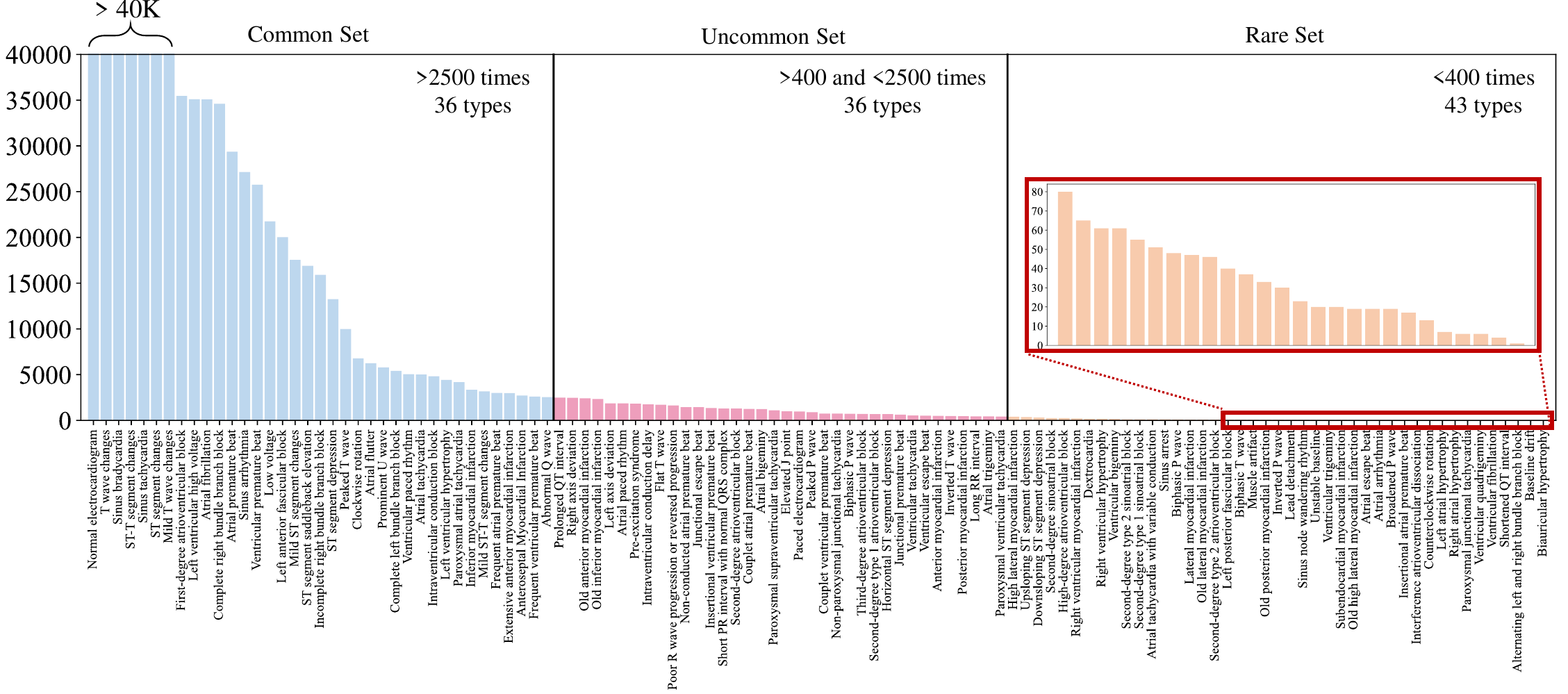}
    \label{fig:datamethod_a}
    \end{subfigure}

    \begin{subfigure}[t]{\textwidth}
    \caption{b. The proposed two-stage ECG diagnosis framework.}
    \includegraphics[width=\linewidth]{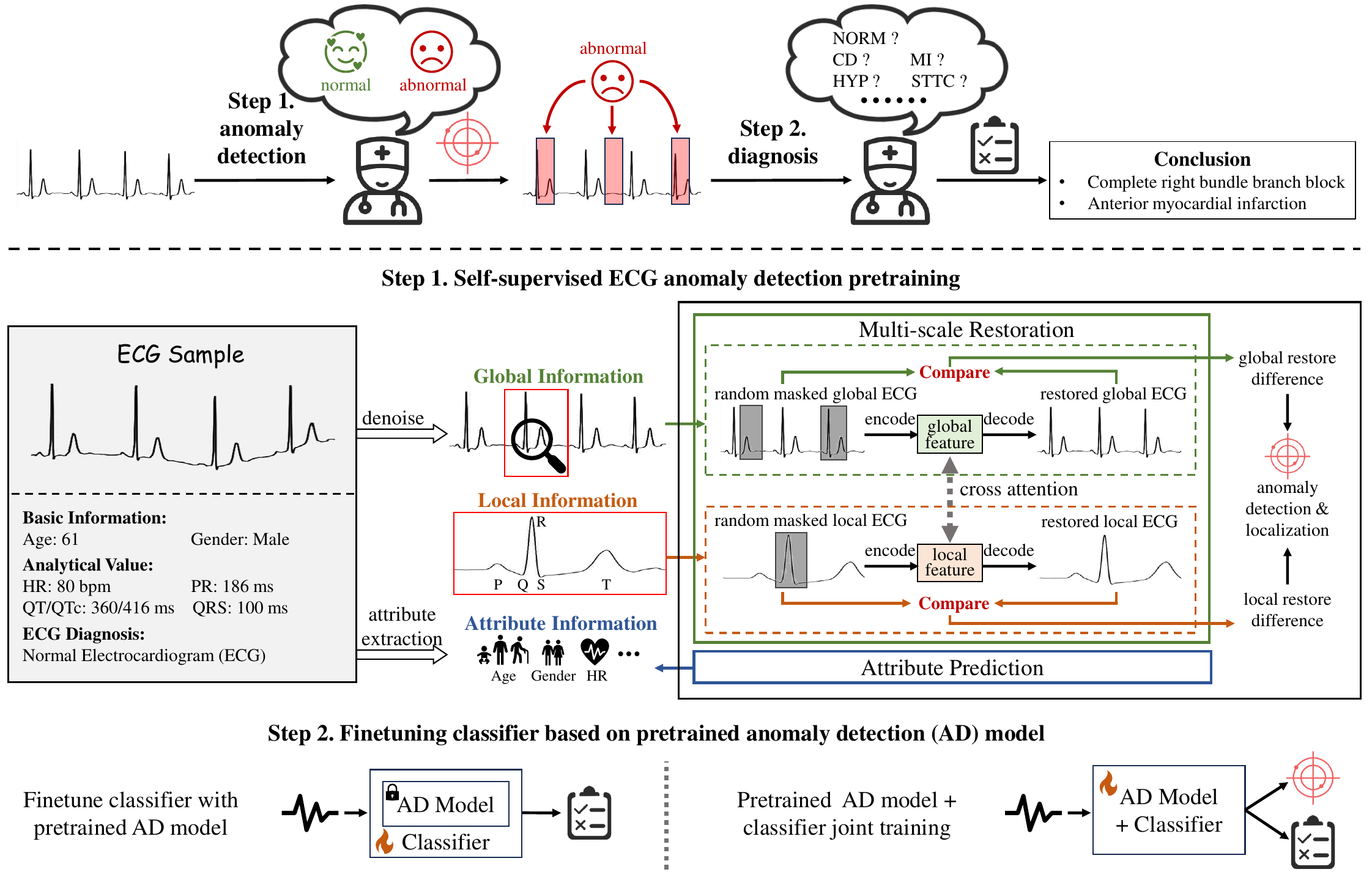}
    \label{fig:datamethod_b}
    \end{subfigure}

    \caption{\textbf{Overview the proposed long-tail ECG diagnosis framework.} 
     (a) Long-tail distribution of cardiac types across the entire dataset. The dataset is characterized by a highly imbalanced distribution, with cardiac types divided into a Common Set, Uncommon Set, and Rare Set based on their frequency of occurrence. The red box highlights the expanded view of the Rare Set, where cardiac types occur fewer than 400 times. (b) The proposed two-stage ECG diagnosis framework. The framework consists of two main steps: Step 1: Self-supervised ECG anomaly detection pretraining, where an anomaly detection model is trained to identify abnormal ECG patterns using both global and local ECG information; and Step 2: Fine-tuning the classifier based on the pre-trained anomaly detection model to provide a detailed diagnosis. This approach improves the classification performance, especially for less common cardiac conditions, by leveraging anomaly detection as pretraining.}
    \label{fig:datamethod}
\end{figure*}

A significant challenge in ECG analysis lies in the long-tail distribution of data (Fig.~\ref{fig:datamethod_a}), where current AI-driven approaches tend to focus on common heart disease patterns, often at the expense of rare or atypical anomalies~\cite{wang2023arrhythmia, rahul2022automatic, nankani2022atrial}. These include critical arrhythmias such as paroxysmal supraventricular tachycardia, ventricular fibrillation, and advanced atrioventricular blocks, which are frequently the precursors to life-threatening cardiovascular events like cardiogenic shock and sudden death. Failure to accurately identify these anomalies can result in missed diagnoses of emergent cardiac conditions~\cite{wuclinical, takayaoutcomes}, underscoring the necessity for an AI model capable of fine-grained, precise classification across a broad spectrum of ECG patterns.

To address the above challenges, we introduce a novel two-stage framework designed to redefine ECG diagnosis as a fine-grained, long-tail classification problem. Our approach begins with anomaly detection pretraining, followed by classification fine-tuning (Fig.~\ref{fig:datamethod_b}). This strategy is rooted in the hypothesis that anomaly detection~\cite{chandola2009anomaly, pang2021deep, fernando2021deep, ukil2016iot} pretraining enhances classification accuracy by first pinpointing anomalous regions within the ECG, making the subsequent classification task more focused and precise. Moreover, by focusing on distinguishing between normal and abnormal signals during the pretraining phase, the model is expect to  more effectively capture the feature of rare anomalies, thereby improving performance on long-tail data distributions. The classification component is seamlessly integrated into the pre-trained anomaly detection model as an additional head, ensuring a unified diagnostic pipeline that mimics the thorough, stepwise analysis conducted by expert cardiologists.

A key innovation of our framework is the incorporation of a self-supervised learning model for anomaly detection pretraining, featuring a novel masking and restoration technique specifically tailored for ECG signal analysis. Central to this framework is a multi-scale cross-attention module that significantly enhances the model’s ability to integrate both global and local signal features. Unlike existing anomaly detection methods that primarily focus on time-series analysis~\cite{li2020survey, venkatesan2018ecg,shen2021time,moody2001impact,wagner2020ptb}, our approach also incorporates essential ECG parameters—such as QRS and QT intervals—along with demographic factors like age and gender, which are crucial for understanding individual heart conditions~\cite{rasmussenprinterval, alqtinterval, doi:10.1161/CIRCEP.119.007284, moss2010gender}. This comprehensive integration provides a more nuanced interpretation of ECG signals, reducing the impact of individual variability and enhancing diagnostic accuracy.

Our methodology was rigorously validated using using the clinical ECG dataset \textbf{ECG-LT}, comprising over one million ECG samples collected from authentic hospital environments in Shanghai between 2012 and 2021. This dataset, which includes \textbf{116} distinct ECG types classified into disease categories, non-specific features, and signal acquisition, allowed us to demonstrate the superior performance of our approach in both internal and external evaluations of ECG diagnosis and anomaly detection/localization. Notably, our approach achieved a 94.7\% AUROC, 92.2\% sensitivity, and 92.5\% specificity for rare ECG types, significantly outperforming existing methods and narrowing the performance gap with common ECG types. In prospective validation studies, cardiologists using our model achieved a 6.7\% improvement in accuracy, an 11.8\% increase in diagnostic completeness, and a 32\% reduction in diagnosis time compared to those working independently. These results suggest the great promise of integrating anomaly detection pretraining into ECG analysis to address the long-standing challenge of long-tail data distributions in clinical diagnostics.

In summary, our work addresses a critical gap in the field of ECG analysis by proposing an innovative approach that significantly improves diagnostic accuracy for both common and rare cardiac conditions. This contribution not only advances the capabilities of AI in healthcare but also has the potential to fundamentally reshape how cardiac diagnostics are performed, ultimately improving patient outcomes on a global scale.

\section{Results}

This section starts with presenting the internal validation findings regarding ECG-LT dataset, including diagnosis performance (Tab.~\ref{tab:class_ruijin}), diagnosis fairness (Fig.~\ref{fig:class_ruijin}) and anomaly detection performance (Tab.~\ref{tab:ad_ruijin}). For external validation, the public benchmark PTB-XL is adopted to evaluate the model's generalizability  (Tab.~\ref{tab:class_ruijin}), with data disparities depicted in Extended Fig.~\ref{fig:datainfo}. Finally, we conducted a rigorous prospective validation by deploying the model in a real-world clinic setting for diagnostic assistance. The comparison of diagnostic accuracy, efficiency, and completeness between cardiologists and our model is shown in Fig.~\ref{fig:prospective_result}.

\subsection{Internal validation on ECG-LT}

\begin{table}[t]
 \caption{Results of ECG diagnosis across multiple evaluation datasets, including internal validation on ECG-LT and external validation on PTB-XL. The best-performing method is highlighted in \textbf{bold}, while the second-best method is \underline{underlined}.}
\centering
\setlength{\tabcolsep}{1.3pt}{
\begin{tabular}{C{1.8cm}|C{1.8cm}|C{1.8cm}C{2.0cm}C{2.5cm}C{3.0cm}C{3.0cm}} 
\toprule
Dataset & \makecell{Evaluation\\ Data} & \makecell{Number of\\ ECGs} & Metrics & \makecell{Supervised\\Classification} & \makecell{Fixed AD$^*$ Model\\ +Classification\\ Finetuning} & \makecell{AD$^*$ Model + \\Classifier\\ Joint Training}\\
\hline
\multirow{12}{*}{\makecell{Internal\\ Validation\\ (ECG-LT)}} & \multirow{3}{*}{\makecell{All\\ Test Data}} & \multirow{3}{*}{189,440} & AUROC & 0.910 & \underline{0.948} & \textbf{0.964}\\
& & & Sensitivity & 0.879 & \underline{0.925} & \textbf{0.931}\\
& & & Specificity & 0.830 & \underline{0.889} & \textbf{0.925}\\
\Xcline{2-7}{0.3pt}
& \multirow{3}{*}{\makecell{Common\\ Test Data}} & \multirow{3}{*}{183,810} & AUROC & 0.940 & \underline{0.962} & \textbf{0.969}\\
& & & Sensitivity & 0.887 & \underline{0.919} & \textbf{0.935}\\
& & & Specificity & 0.864 & \underline{0.905} & \textbf{0.913}\\
\Xcline{2-7}{0.3pt}
& \multirow{3}{*}{\makecell{Uncommon\\ Test Data}} & \multirow{3}{*}{5,253} & AUROC & 0.932 & \underline{0.968} & \textbf{0.977}\\
& & & Sensitivity & 0.872 & \underline{0.915} & \textbf{0.937}\\
& & & Specificity & 0.865 & \underline{0.921} & \textbf{0.938}\\
\Xcline{2-7}{0.3pt}
& \multirow{3}{*}{\makecell{Rare\\ Test Data}} & \multirow{3}{*}{386} & AUROC & 0.858 & \underline{0.914} & \textbf{0.947}\\
& & & Sensitivity & 0.876 & \underline{0.877} & \textbf{0.922}\\
& & & Specificity & 0.762 & \underline{0.912} & \textbf{0.925}\\
\hline
\multirow{3}{*}{\makecell{External\\ Validation\\ (PTB-XL)}} & \multirow{3}{*}{\makecell{External\\ Test Data}} & \multirow{3}{*}{2,198} & AUROC & 0.858 & \underline{0.876} & \textbf{0.896}\\
& & & Sensitivity & 0.774 & \underline{0.818} & \textbf{0.835}\\
& & & Specificity & 0.790 & \underline{0.805} & \textbf{0.848}\\
\bottomrule

\multicolumn{4}{l}{\small *AD: Anomaly Detection.}
\end{tabular}
}
\label{tab:class_ruijin}
\end{table}

\subsubsection{ECG Diagnosis}
To thoroughly evaluate diagnostic performance in scenarios with long-tail distributions presented in Fig.~\ref{fig:datamethod_a}, we divided the test dataset into three subsets based on ECG type frequency: common, uncommon, and rare. As illustrated in Tab.~\ref{tab:class_ruijin}, performance markedly declines from common to rare ECG types when using a straightforward supervised classification approach, with the AUROC decreasing from 94.0\% for common data to 85.8\% for rare data. Incorporating a pretrained anomaly detection model trained on normal ECG data significantly mitigates this performance drop. Both settings, 1)  fixing the anomaly detection model and simply fine-tuning the classifier and 2) further jointly training the anomaly detection model and classifier, demonstrate improved metrics across all subsets and with more substantial enhancements in handling long-tail rare data. 

When the anomaly detection model was further jointly trained with the classifier, the overall classification performance metrics reached new heights, with an AUROC of 96.4\%, sensitivity of 93.1\%, and specificity of 92.5\% across all test data. For rare ECG types, the results were particularly compelling, with the evaluation metrics achieving a 94.7\% AUROC, 92.2\% sensitivity, and 92.5\% specificity. The specific AUROC performance for each ECG type within the rare test set is detailed in Fig.~\ref{fig:class_ruijin_a}. Notably, the joint training of the anomaly detection model and classifier achieved an AUROC exceeding 95\% for most anomaly types, underscoring the significant advantage of this approach in long-tail classification scenarios.

These findings provide robust evidence for the critical role that anomaly detection models play in enhancing diagnostic accuracy, particularly in the context of rare, underrepresented ECG types. The results highlight the effectiveness of our approach in addressing the inherent challenges of long-tail data distributions in clinical diagnostics.

\subsubsection{Diagnosis fairness}
Beyond assessing overall diagnostic performance, it is essential to ensure fairness across key demographic characteristics, especially given the clinical need for consistent accuracy across diverse age groups and genders. Our analysis reveals comparable diagnostic performance between males and females, as shown in Fig.~\ref{fig:class_ruijin_b}.
Males exhibit slightly better diagnostic performance, with an increase of 0.8\% in AUROC, 1.2\% in sensitivity, and 1.7\% in specificity, respectively.
For different age groups across all test data, as shown in Fig.~\ref{fig:class_ruijin_c}, the diagnosis performance shows a lower AUROC of 88\% and Specificity of 78\% for patients under 10 years old, as well as a Specificity of 85\% for patients over 90 years old. In contrast, other age groups between 10 and 90 years old maintain similar AUROC and Specificity values exceeding 90\%.

\subsubsection{Anomaly detection and localization}

Tab.~\ref{tab:ad_ruijin} provides a comprehensive evaluation of anomaly detection methods on test sets with varying degrees of anomaly rarity. Our method consistently outperforms others in most evaluation metrics across all test datasets, achieving an AUROC of 91.2\%, an F1 score of 83.7\%, a sensitivity of 84.2\%, a specificity of 83.0\%, and a precision of 75.6\% at a fixed 90\% recall. These results significantly surpass those of competing methods, with gains of 8.8\% in AUROC, 9.1\% in F1 score, and 14.3\% in precision. While methods like TranAD and AnoTran exhibit nearly perfect sensitivity, their specificity scores are lower, indicating challenges in accurately detecting anomalies. Our methodology also demonstrates exceptional effectiveness across varying levels of anomaly rarity, with AUROC scores of 91.4\%, 89.5\%, and 89.6\% for the common, uncommon, and rare test sets, respectively, underscoring our method’s robustness in detecting rare anomalies.

\begin{figure*}[p]
\centering
    \centering
    \begin{subfigure}[t]{\textwidth}
    \centering
    \vspace{-1.2cm}
    \subcaptionbox{a. Diagnosis performance on the tail classes\label{fig:class_ruijin_a}}
    {
        \includegraphics[width=0.9\linewidth,]{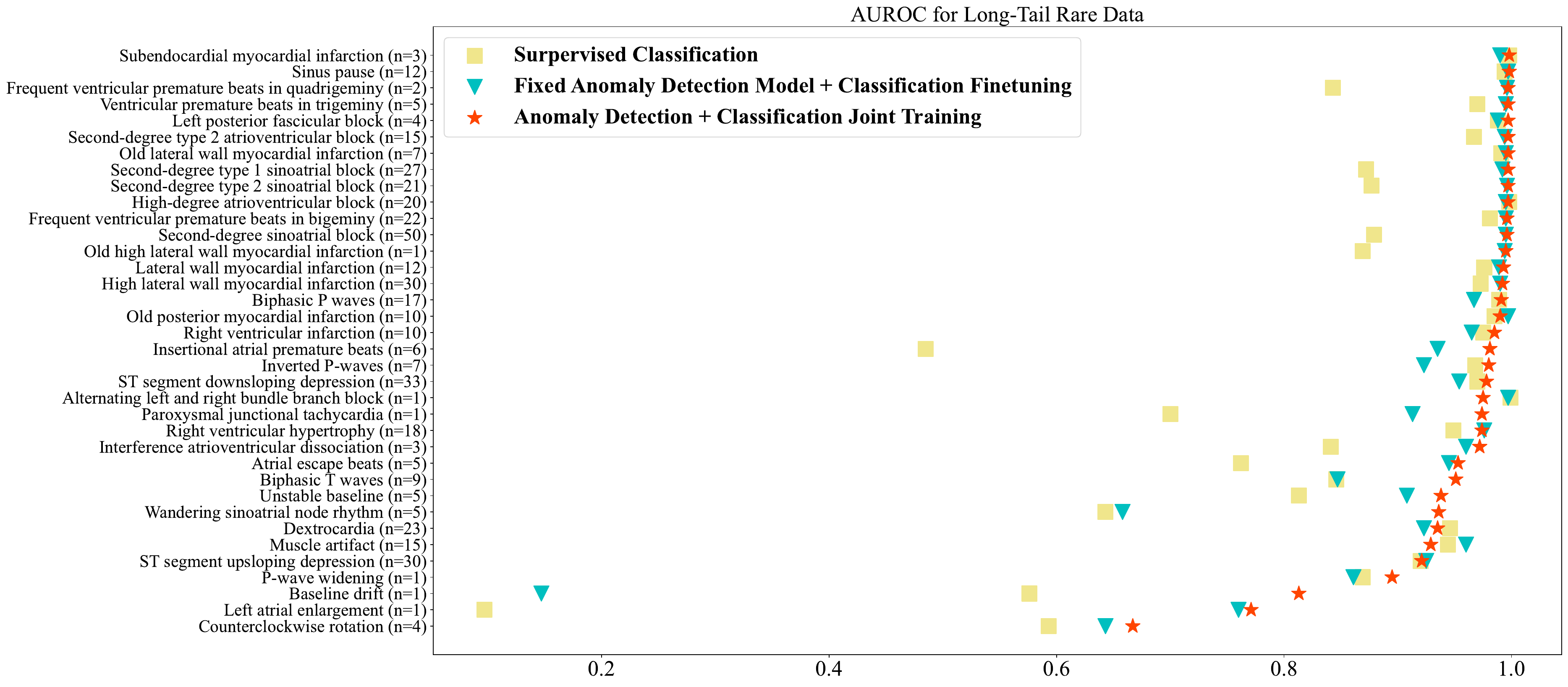}
    }
    \end{subfigure}

    \begin{subfigure}[t]{\textwidth}
    \centering
    \vspace{-0.2cm}
    \subcaptionbox{b. Diagnosis fairness across sex\label{fig:class_ruijin_b}}
    {
        \includegraphics[width=0.9\linewidth]{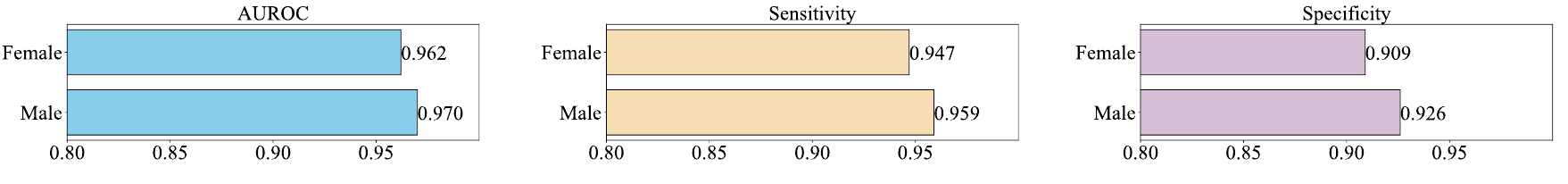}
    }
    \end{subfigure}

    \begin{subfigure}[t]{\textwidth}
    \centering
    \vspace{-0.2cm}
    \subcaptionbox{c. Diagnosis fairness across age\label{fig:class_ruijin_c}}
    {
        \includegraphics[width=0.9\linewidth]{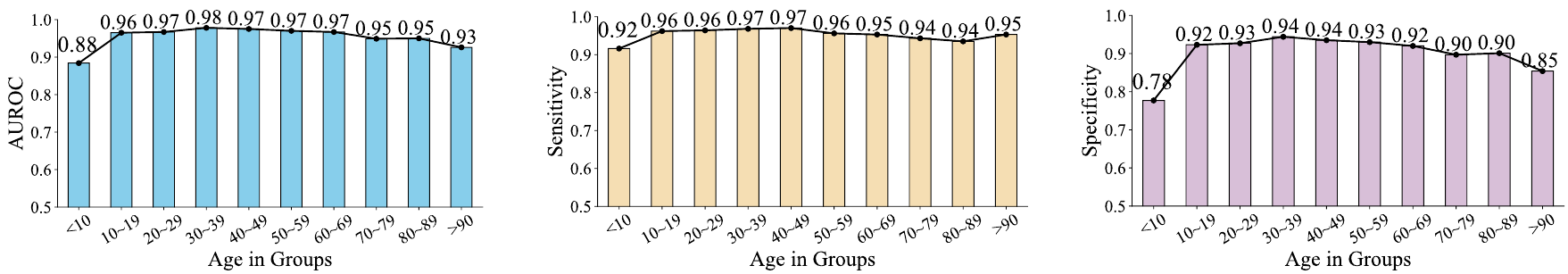}
    }
    \end{subfigure}

    \begin{subfigure}[t]{\textwidth}
    \centering
    \vspace{-0.2cm}
    \subcaptionbox{d. Visualization of anomaly localization\label{fig:class_ruijin_d}}
    {
        \includegraphics[width=0.9\linewidth]{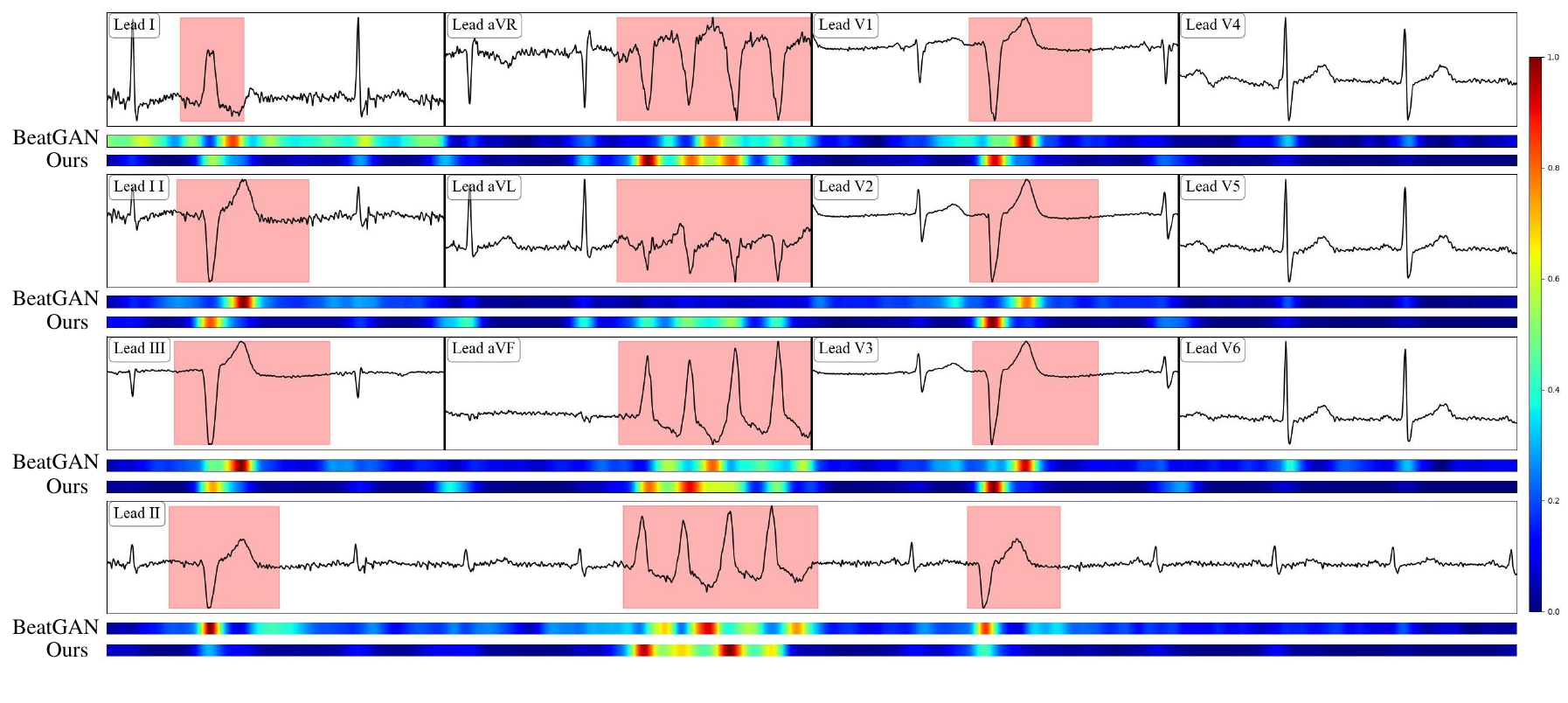}
    }
    
    \end{subfigure}

\caption{\textbf{Performance of ECG diagnosis.} 
(a) Diagnosis performance on the tail classes. A comparison of ECG diagnosis for each type in the Rare Set. The proposed method demonstrates superior performance, especially for anomalies with fewer samples.
(b) Diagnosis fairness across sex. The model maintains consistent performance across male and female subjects, demonstrating balanced accuracy between sexes.
(c) Diagnosis fairness across age. Diagnosis performance is  presented for different age groups in ten-year intervals. The model exhibits stable and equitable performance across all age groups.
(d) Visualization of anomaly localization. The proposed method's localization (Ours) is compared with a leading baseline (BeatGAN), with ground truth marked in pink boxes. Color-coded scores (ranging from 0 to 1) indicate the likelihood of an anomaly, with red marking the most likely location of anomaly.}
\label{fig:class_ruijin}
\end{figure*}

In the cardiologist-annotated localization test set, our method demonstrates significant improvements, achieving the highest results with a Dice coefficient of 65.3\% and an AUROC of 76.5\%, as detailed in Tab.~\ref{tab:ad_ruijin}. These results represent a substantial a substantial advancement over the closest rival, BeatGAN, with improvements of 7.3\% in the Dice score and 11.6\% in AUROC. Methods like TranAD and AnoTran, which struggle to account for individual variances and face challenges in learning from large-scale ECG datasets, exhibited reduced effectiveness in anomaly localization. 

For a comprehensive visual analysis of anomaly localization, Fig.~\ref{fig:class_ruijin_d} compares the performance of our method against BeatGAN using a 12-lead ECG test example. The anomaly regions, highlighted in red by the expert cardiologists, serve as the ground truth. Our approach effectively emphasizes critical anomaly areas, such as those in Lead aVL, while minimizing false positives in areas identified as normal, such as Lead I. 
Our method demonstrates a heightened sensitivity to abrupt signal pattern changes at a granular level, resulting in more precise anomaly localization compared to the broader annotations provided by the ground truth. These precise localizations offer valuable insights into potential anomalies, providing significant support for medical practitioners. This effectiveness is further validated by assessments from experienced cardiologists.

\begin{table}[t]
 \caption{Anomaly detection and localization results across multiple evaluation data, compared to state-of-the-art methods. The most effective method is highlighted in \textbf{bold}, and the second-best is \underline{underlined}.
 }
\centering
\scalebox{0.95}{
\begin{tabular}{C{1.7cm}C{2.3cm}C{1.5cm}|C{1.8cm}C{1.8cm}C{1.8cm}C{1.8cm}|C{1.2cm}} 
\toprule
\multirow{2}{*}{Task} & \multirow{2}{*}{Evaluation Data} & \multirow{2}{*}{Metrics} & \multicolumn{3}{r}{State-of-the-art Methods} & & \multirow{2}{*}{Ours}\\
 &  &  & TranAD~\cite{tranad} & AnoTran~\cite{xu2022anomaly} & BeatGAN~\cite{liu2022time} & TSL~\cite{zheng2022task} & \\
\Xcline{1-8}{0.3pt}
\multirow{20}{*}{\makecell[c]{Anomaly \\Detection}} & \multirow{5}{*}{\makecell{All\\ Test Data}} & AUROC  & 0.566  &  0.601   & 0.653   & \underline{0.824} & \textbf{0.912}\\
& &  F1 Score & 0.667  &  0.673   & 0.669  & \underline{0.746} & \textbf{0.837}\\
& & Sensitivity & 0.511  &  0.579   & 0.522  & \underline{0.793} & \textbf{0.842}\\
& & Specificity & 0.514  &  0.564  & \underline{0.698}   & 0.668 & \textbf{0.830}\\
& & Pre@90   & 0.501  &  0.530   & 0.528  & \underline{0.613} & \textbf{0.756} \\
\Xcline{2-8}{0.3pt}
& \multirow{5}{*}{\makecell{Common\\ Test Data}} & AUROC & 0.565  &  0.599   & 0.650   & \underline{0.820} & \textbf{0.914}\\
& & F1 Score  & 0.667  &  0.672   & 0.669   & \underline{0.744} & \textbf{0.839}\\
& & Sensitivity  & 0.512  &  0.576  & 0.537   & \underline{0.792} & \textbf{0.852}\\
& & Specificity  & 0.514  &  0.564   & \underline{0.679}   & 0.662 & \textbf{0.820}\\
& & Pre@90   & 0.501 &  0.529  & 0.528  & \underline{0.611} & \textbf{0.761} \\
\Xcline{2-8}{0.3pt}
& \multirow{5}{*}{\makecell{Uncommon\\ Test Data}} & AUROC  & 0.578  &  0.623   & 0.681   & \underline{0.853} & \textbf{0.895}\\
& & F1 Score  & 0.667 &  0.677   & 0.675   & \underline{0.773} & \textbf{0.819}\\
& & Sensitivity  & 0.504  & 0.628  & 0.552   & \underline{0.753} & \textbf{0.814}\\
& & Specificity  & 0.523 & 0.547   & 0.719   & \underline{0.805} & \textbf{0.828}\\
& & Pre@90   & 0.502  &  0.536   & 0.536   & \underline{0.625} & \textbf{0.700}\\
\Xcline{2-8}{0.3pt}
& \multirow{5}{*}{\makecell{Rare\\ Test Data}} & AUROC  & 0.568  & 0.604   & 0.659  & \underline{0.888} & \textbf{0.896}\\
& & F1 Score  & 0.667  &  0.676  & 0.673   & \underline{0.816} & \textbf{0.827}\\
& & Sensitivity  & 0.513 &  0.582  & 0.510   & \underline{0.769} & \textbf{0.825}\\
& & Specificity  & 0.502  &  0.577   & 0.746  & \textbf{0.885} & \underline{0.831}\\
& & Pre@90   & 0.496  &  0.531   & 0.521  & \textbf{0.693} & \underline{0.692} \\
\hline
\multirow{2}{*}{\makecell[c]{Anomaly \\Localization}} & \multirow{2}{*}{\makecell[c]{Localization\\ Test Set}} & AUROC  & 0.602  & 0.525   & \underline{0.649}  & 0.537 & \textbf{0.765}\\
& & Dice  & 0.549  &  0.511  & \underline{0.580}   & 0.567 & \textbf{0.653}\\
\bottomrule
\end{tabular}
}
\label{tab:ad_ruijin}
\end{table}

\subsection{External validation on public benchmarks}

In Tab.~\ref{tab:class_ruijin}, we present the results of external validation for our method and baseline models using the publicly available PTB-XL benchmark dataset. This dataset differs from our proprietary ECG-LT dataset in terms of age distribution, signal acquisition quality, and types of ECG signals. The distinct characteristics of the internal and external validation datasets are visually depicted in Extended Fig.~\ref{fig:datainfo}. When applying our method to the external validation dataset through linear probing, the jointly trained anomaly detection model and classifier achieve the highest diagnostic accuracy, with an AUROC of 89.6\%, sensitivity of 83.5\%, and specificity of 84.8\%. This performance surpasses the other two baselines, with AUROC improvements of 3.8\% and 2.0\%, respectively. Notably, only the final linear layer of the classifier is involved in the linear probing process, while the remaining model parameters remain fixed. All three competing methods undergo identical linear probing procedures.

\subsection{Prospective validation}
To rigorously evaluate the model's performance in a real-world clinical setting, we deployed it in a hospital environment to assist cardiologists with diagnoses, without any fine-tuning. Our prospective validation was conducted from May to July 2024, utilizing data from the emergency department at Ruijin Hospital in Shanghai, China. Detailed information is provided in Extended Tab.~\ref{tab:dataset_cohort}. Each ECG was evaluated by at least three cardiologists under different conditions: 
\begin{itemize}
 \item Cardiologist A was tasked with providing conclusions as quickly as possible, simulating emergency scenarios where rapid decision-making is crucial. 
 \item Cardiologist B provided conclusions independently without time constraints, representing a routine diagnostic process. 
\item Cardiologist C made diagnoses with the assistance of our deployed model, which offered the five most likely ECG types for each case as references. 
\end{itemize}
Diagnosis times were recorded without the cardiologists' knowledge to ensure an unbiased assessment of efficiency. Additionally, an adjudicating cardiologist with the highest level of expertise reviewed and confirmed the diagnoses provided by these clinicians to ensure accuracy and completeness.

\begin{figure}[t]
\centering
\includegraphics[width=1.0\textwidth]{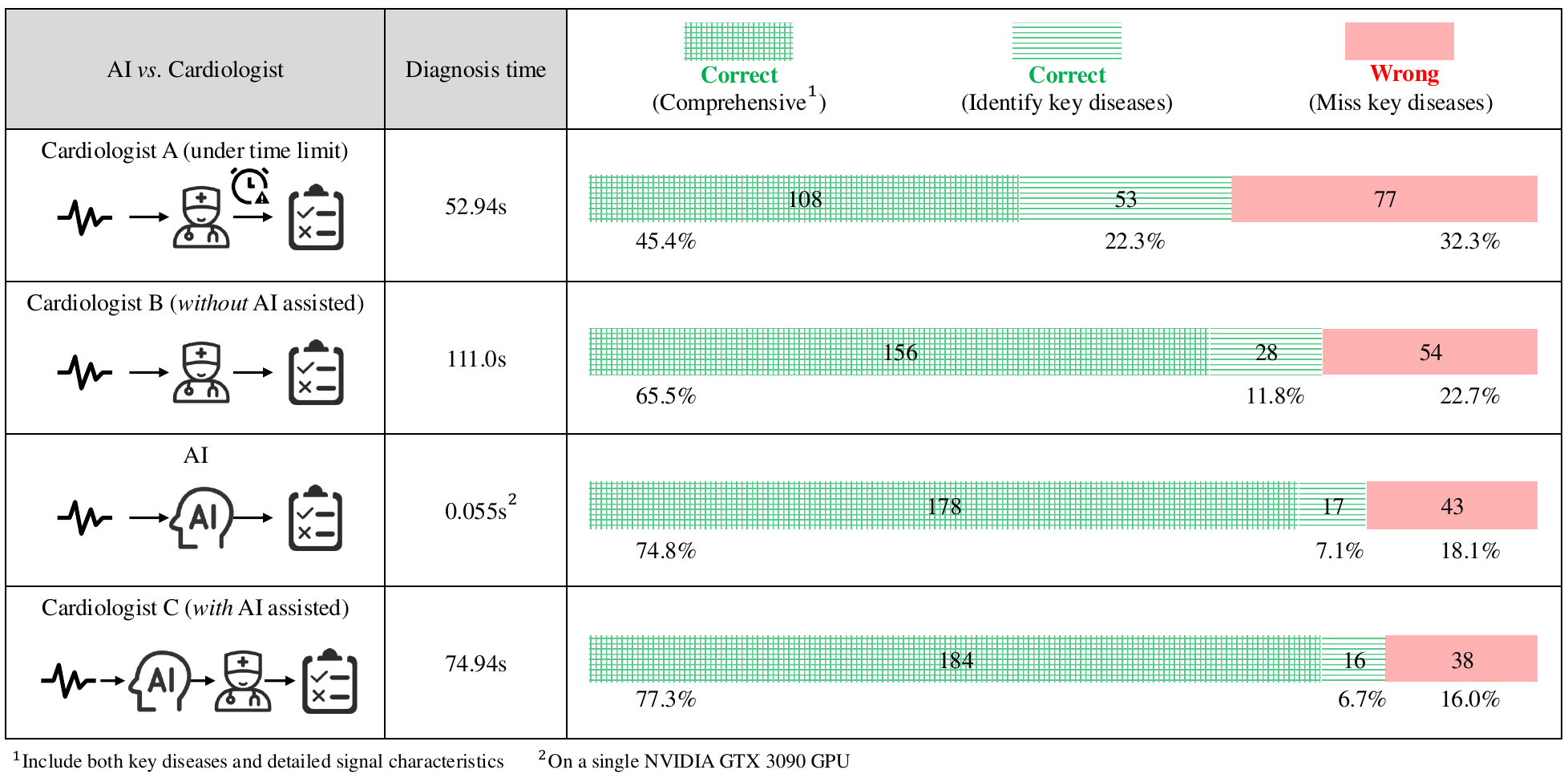}
\caption{
Prospective validation results from the emergency department at Ruijin Hospital, comparing diagnostic timing, completeness, and accuracy between cardiologists and an AI-assisted system. The figure presents the diagnosis time and accuracy rates for Cardiologist A (under time limit), Cardiologist B (without AI assistance), AI alone, and Cardiologist C (with AI assistance). The metrics include the correct identification of comprehensive diagnoses (key diseases and detailed signal characteristics) and the correct identification of key diseases, along with the frequency of missed key diseases. The AI-assisted diagnosis demonstrated superior performance in both speed and accuracy, significantly reducing diagnostic time compared to cardiologists, especially under time constraints.}
\label{fig:prospective_result}
\end{figure}

\begin{table}[t]
 \caption{Prospective validation results from the Emergency Department at Ruijin Hospital: An analysis of sensitivity and specificity for both common and long-tail anomalies.}
\centering
\setlength{\tabcolsep}{1.3pt}{
\begin{tabular}{C{1.8cm}|C{1.8cm}|C{3.0cm}C{3.2cm}C{3.0cm}C{3.0cm}} 
\toprule
Evaluation Data & Metrics & Cardiologist A (under time limit) & Cardiologist B (\textit{without} AI assisted) & AI & Cardiologist C (\textit{with} AI assisted)\\
\hline
\multirow{2}{*}{\makecell{Common\\ data}} & Sensitivity  & 0.626 & 0.747 & 0.837 & 0.820\\
& Specificity & 0.992 & 0.973 & 0.916 & 0.989\\
\hline
\multirow{2}{*}{\makecell{Long-tail\\ data}} & Sensitivity  & 0.347 & 0.469 & 0.592 & 0.714\\
& Specificity & 0.997 & 0.997 & 0.986 & 0.997\\
\bottomrule
\end{tabular}
}
\label{tab:prospective_result}
\end{table}

As shown in Fig.~\ref{fig:prospective_result}, under time constraints, cardiologists demonstrated lower accuracy, as reflected by Cardiologist A's performance, which was significantly impacted, resulting in less comprehensive conclusions that primarily focused on identifying key diseases, with only 45.4\% of diagnoses being comprehensive and 32.3\% missing key diseases. In contrast, without time limits, Cardiologist B achieved a comprehensive diagnosis rate of 65.5\%, missing key diseases in only 22.7\% of cases. Our AI method analyzes an ECG instance in just 0.055 seconds, approximately 1,000 times faster than the average time required for human emergency diagnosis. Beyond its speed, the AI method achieves a higher diagnostic accuracy rate of 81.9\%, outperforming the 67.7\% accuracy rate for unaided human diagnoses. When integrated into clinical practice, AI-assisted cardiologists, as exemplified by Cardiologist C, attained an accuracy rate of 84.0\%, reflecting a 6.7\% improvement over unaided diagnoses. Additionally, diagnostic efficiency significantly improved, with an average reduction of 36 seconds in diagnosis time. The AI system also provided more detailed insights into signal patterns and rhythms, contributing to more comprehensive conclusions in 11.8\% of ECGs, particularly in recognizing subtle changes like T wave alterations and sinus tachycardia, thereby enhancing the quality of diagnostic outcomes.

In clinical diagnosis, particularly for long-tail anomalies, cardiologists under time constraints or with less experience are prone to misdiagnosis, often demonstrating high specificity (>99\%) but very low sensitivity (<50\%). 
Integrating AI into the diagnostic process significantly mitigates these errors by enhancing the detection of rare anomalies and highlighting critical signal patterns.
As illustrated in Tab.~\ref{tab:prospective_result}, our model, when used as an assistive tool, improved the sensitivity of cardiologists on long-tail data from 46.9\% to 71.4\%, while maintaining a high specificity of 99.7\%. This demonstrates the potential of AI to augment clinical decision-making, especially in challenging diagnostic scenarios.

\section{Discussion}

\noindent \textbf{Significant potential for clinical application.}
Our ECG diagnosis model, utilizing anomaly detection pretraining, achieves accurate and comprehensive results significantly faster than experienced cardiologists. This highlights the potential of an AI-assisted system in clinical diagnosis, applicable to both emergency and routine ECG evaluations.  In high-pressure situations, where errors or omissions of critical features can occur, our method can help mitigate these risks. As shown in Fig~\ref{fig:prospective_result}, the combination of greatly reduced diagnostic time and high accuracy, thoroughly validated through clinical trials showing superior performance to human diagnosis under time constraints, points to a promising application in emergency settings. This advancement is particularly crucial for optimizing the critical golden period in cases of severe and rapidly progressing heart disease. For routine diagnoses, our model not only surpasses the accuracy of experienced cardiologists but also provides more detailed signal information, identifying all potential diseases. Rigorous evaluations and statistical analyses reveal that even less experienced cardiologists, when assisted by our model, can reach more accurate and comprehensive conclusions than those made by experienced cardiologists alone, suggesting the practical capability of our model to help cardiologists avoid overlooking significant conditions.

\noindent \textbf{Anomaly detection pretraining mitigates the influence of long-tail distributions.}
In ECG diagnosis, a significant challenge in developing AI-powered analysis systems arises from the long-tail distribution of numerous anomaly types, where different anomalies occur with vastly unequal frequencies. This imbalance has considerably hindered progress in ECG diagnosis model development, causing most current models to focus primarily on detecting common anomalies. To address this issue, we propose a self-supervised anomaly detection approach as a pretraining method.
This approach identifies deviations corresponding to potential abnormal features, enabling the model to classify anomaly types with greater precision by concentrating on specific abnormal regions. This heightened sensitivity to anomalous regions facilitates more efficient learning across various levels of anomaly rarity, thereby mitigating the challenges associated with imbalanced long-tail anomaly distributions.
By integrating anomaly detection pretraining into the supervised classification process, our model develops a more comprehensive understanding of  diverse ECG anomalies, leading to enhanced diagnostic performance and versatility, particularly in accurately identifying rare anomalies.

\noindent \textbf{Anomaly detection pretraining yields explainable and informative localization results.}
Beyond its enhanced diagnostic performance, anomaly detection pretraining offers a crucial advantage: the ability to precisely localize anomalies, thereby providing clear and understandable justifications for the model’s diagnostic decisions. Such transparency is crucial for developing an AI-driven diagnostic system that is both trustworthy and convincing. By pinpointing specific areas of anomaly within the ECG signal, these localization results offer cardiologists precise cues, directing their attention to potential regions of concern.
This aids in the accurate identification and interpretation of anomalous patterns indicative of cardiovascular irregularities, significantly enhancing the practical application of such systems in real hospital diagnostic scenarios.

\noindent \textbf{Fairness of the clinical diagnosis models.}
Ensuring unbiased clinical diagnosis requires not only high overall diagnostic accuracy but also fairness across key demographic variables, with particular attention to age and gender. As shown in Fig.~\ref{fig:class_ruijin_b} and Fig.~\ref{fig:class_ruijin_c}, our analysis demonstrates comparable diagnostic efficacy between males and females, as well as across various age groups from 10 to 90 years. However, a slight decrease in diagnostic performance is observed in patients under 10 years old and those over 90, compared to other age groups. This disparity may stem from physiological differences and varied disease presentations at these age extremes. These findings highlight the importance of accounting for demographic factors to enhance both diagnostic accuracy and fairness in clinical practice. Further research is warranted to explore the mechanisms behind these age and gender disparities, as it is crucial to develop strategies to improve healthcare outcomes for all patient populations.

\noindent \textbf{Extensible framework for ECG diagnosis.}
Our framework is specifically designed to address the long-tail distribution of ECG data, having been meticulously trained on the ECG-LT dataset, which encompasses 116 distinct ECG types. This comprehensive coverage ensures that the model is capable of accommodating nearly all ECG types encountered in clinical practice, making it highly adaptable to datasets with diverse distributions. By simply fine-tuning the final classification layer, the framework trained on the ECG-LT dataset demonstrates superior diagnostic performance on external ECG datasets, as illustrated in Tab.~\ref{tab:class_ruijin}. This capability underscores the model’s proficiency in capturing the inherent complexity of ECG signals, making it versatile across different data sources. As a result, this evidence strongly supports the framework's generalizability and practical utility in ECG diagnosis.

\noindent \textbf{Component analysis.}
This study focuses on the development and evaluation of a computer-aided ECG diagnosis system based on a novel self-supervised anomaly detection pretraining framework, which significantly enhances ECG classification performance in cases with long-tail distribution. The system's impressive performance is attributed to three innovative design elements:
1) A unique masking and restoring technique that enables the model to learn ECG characteristics by interacting with adjacent unmasked areas;
2) The implementation of multi-scale ECG analysis, which, when cross-referenced, mimics the analytical approach of experienced cardiologists;
3) The incorporation of ECG parameters as well as demographic information, which has not been previously considered in such contexts.
These critical design elements are embodied within four main components of our framework: the Masking and Restoring (MR) module, the Multi-Scale Cross-Attention (MC) module, the Trend Assisted Restoration (TAR), and the Attribute Prediction Module (APM). The ablation study starting with a basic model that analyzes global ECG patterns and progressively introducing these additional components, as shown in Extended Fig.~\ref{fig:abl_ruijin}, demonstrates the positive impact of each component on the overall effectiveness of the system.

\noindent \textbf{Attribute prediction.} To enhance the model's capability to capture the association between ECG signals and patient attributes, the Attribute Prediction Module (APM) enables the model to predict attributes for new ECG samples, aiding in anomaly detection and diagnosis. This module provides valuable insights into the model's understanding of ECG data. The attributes involved encompass patient demographics such as age and gender, as well as key physiological indicators like heart rate and PR interval. The attribute prediction results, illustrated in Extend Tab.~\ref{tab:abl_apvalue} and Extend Fig.~\ref{fig:abl_apm}, demonstrate the module’s strong performance across both normal and abnormal data, with slightly higher accuracy observed for normal instances. For instance, gender prediction accuracy decreases from 86.5\% in normal ECGs to 75.9\% in abnormal ECGs. A similar trend is observed in numerical attributes, where normal ECGs tend to exhibit smaller deviations. This observation aligns with the fact that the model is exclusive pretraining on normal data, reaffirming its ability to accurately discern the relationship between signals and attributes.

\noindent \textbf{Clinical value.} Our research incorporates fundamental heart rhythm analysis, akin to the "first glance" of a human cardiologist at an electrocardiogram. This approach is particularly relevant as China has recently made significant improvements in treating cardiovascular diseases through the strategic redistribution of medical resources. However, there remains a pressing need for effective monitoring models in primary healthcare settings. These models must empower non-specialist healthcare providers to accurately assess abnormal ECG signals, regardless of whether they are using low-cost or high-end equipment. Such capability is critical for early detection and prevention of acute cardiovascular events, underscoring the importance of our research.

\noindent \textbf{Future work.} This study emphasizes the clinical utility of the typical 10-second 12-lead ECG for evaluating heart rhythm. However, the limitations of short-term ECGs in identifying intermittent cardiac abnormalities underscore the need for extended monitoring. The consensus within the medical community on incorporating long-term ECG data is driven by the potential to uncover novel digital biomarkers and dynamic indicators of cardiovascular health. We are optimistic about expanding our methodology to include more comprehensive datasets due to its inherent scalability. Moving forward, our goal is to  develop advanced algorithms capable of processing and integrating diverse types of data, including pairing ECG data with echocardiography to create a more nuanced diagnostic tool~\cite{bello2015role}. Such an integrated approach not only promises to enhance diagnostic accuracy but also to identify severe disease conditions requiring prompt medical intervention. This effort aims to ensure that patients do not miss the critical window for diagnosis and treatment, ultimately paving the way for significant advancements in the prevention and management of cardiovascular diseases.

\section{Methods}

\subsection{Dataset}

In this section, we describe the datasets used to train the anomaly detection model and downstream classifiers. In Sec.~\ref{sec:dataruijin}, we present detailed information on the internal ECG dataset. To the best of our knowledge, this is not only the largest ECG dataset, containing over one million ECG samples from real-world clinical practice, but it also encompasses the most comprehensive range of cardiac anomaly types, particularly the long-tail ones. In Sec.~\ref{sec:dataopen}, we introduce the open-source datasets used for external validation. 

\subsubsection{ECG-LT: internal dataset}
\label{sec:dataruijin}
We conducted a comprehensive assessment using an extensive dataset obtained from real-world clinical practice, comprising  \textbf{1,089,367 ECG samples} collected between 2012 and 2021. Each sample includes an ECG signal image and a diagnostic summary identifying specific anomalies. This dataset captures a wide spectrum of ECG abnormalities, ranging from common conditions like atrioventricular block to rare conditions like biventricular hypertrophy, spanning 116 unique types. The distribution of these anomalies is illustrated in Fig.~\ref{fig:datamethod_a}, where common cardiac conditions have tens of thousands of samples, while rare conditions have only single-digit instances, highlighting a pronounced long-tail distribution. These 116 types are broadly categorized into disease classification, non-specific features, and signal acquisition, with the hierarchical structure visually presented in Extended Fig.~\ref{fig:datainfo}a. Furthermore, the dataset includes patient demographics such as gender, age, heart rate, and key cardiac intervals like PR, QT, corrected QT, and QRS complex. We transformed these samples into time-series data for 12-lead ECGs, where each lead represents 2.5 seconds of data, except for lead II, which shows a full 10 seconds, all at a sampling rate of 500Hz.

We gathered all ECG recordings up to the year 2020, resulting in a total of 416,951 normal ECGs and 482,976 abnormal ECGs, for  model training. To effectively evaluate the model's classification performance in this long-tail scenario, we used ECGs from the year 2021 for internal validation, comprising  94,304 normal ECGs and 95,136 abnormal ECGs. These validation ECGs were divided into three distinct test sets: \textbf{common}, \textbf{uncommon}, and \textbf{rare}. The common test set includes 37 cardiac types occurring more than 2,500 times, while the rare test set contains 43 cardiac types occurring less than 400 times. The remaining 36 cardiac types are categorized as uncommon. The distribution of age and gender across both the training and internal validation sets was consistent, as shown in Extended Fig.~\ref{fig:datainfo}c and Extended Fig.~\ref{fig:datainfo}d. Furthermore, to create a test set with precise anomaly localization annotations at the level of individual signal points, two experienced cardiologists manually annotated the exact locations of anomalies on 500 ECGs. 

\subsubsection{External open-source dataset}
\label{sec:dataopen}

The PTB-XL~\cite{wagner2020ptb} database includes 21,837 clinical 12-lead ECGs obtained from 18,885 patients, each lasting 10 seconds and sampled at a rate of 500Hz. These recordings were captured using devices manufactured by Schiller AG, spanning from October 1989 to June 1996. Within the PTB-XL database, there are 23 distinct subclasses of ECG recordings, 18 of which overlap with our internal ECG-LT dataset, as shown in Extended Fig.~\ref{fig:datainfo}a and Extended Fig.~\ref{fig:datainfo}b. The left 5 ECG types, exclusive to ECG-LT, arise from the delineation of specific pathological conditions in diagnostic conclusions, an approach not used in the recording protocols of the ECG-LT dataset due to differing methodologies. In line with the PTB-XL dataset's framework, we performed external validation using a split test set comprising 2,198 ECGs. Furthermore, the distribution of age and gender between internal and external validation sets is visually compared in Extended Fig.~\ref{fig:datainfo}c and Extended Fig.~\ref{fig:datainfo}d. External validation sets exhibit 51.5\% male representation, closely matching the 51.7\%  observed in internal validation sets, with a significant proportion of patients aged 70 to 90 years.
\subsection{Long-tail diagnosis model for ECG}

This section introduces the learning framework designed to integrate anomaly detection pretraining into ECG classification. We first provide a detailed overview of the proposed model, followed by its training procedures, which consist of two steps: self-supervised anomaly detection and supervised classification through fine-tuning.

Each ECG training sample comprises two elements: $\mathcal{S} = {\mathcal{X}, \mathcal{T}}$. Here, $\mathcal{X}$ represents the signal components, consisting of $D$ signal points extracted from ECG samples, which serve as the input to our framework. $\mathcal{T}$ encompasses the textual components of ECG samples, including patient-specific information (e.g., age and gender), fundamental attributes (e.g., heart rate, PR interval), and the diagnostic conclusion, which serve as the supervision signals for our framework’s output. The objective is to identify potential anomaly regions and predict possible cardiac anomaly class based on the input ECG signal. The training is achieved through separate self-supervised anomaly detection and supervised classification steps.

\subsubsection{Self-supervise anomaly detection pretraining}
\label{sec:adarch}

The self-supervised anomaly detection pretraining is designed to restore randomly masked segments of ECG signals and forecast the attribute data recorded within the ECG samples. When a new test ECG is presented, the model predicts anomaly scores and generates detailed score maps by comparing the restored signal with the original. Notably, this process does not require diagnostic information during training; instead, only attribute data such as age and gender are used as supervisory signals. As illustrated in Extended Fig.~\ref{fig:model_supp}, the self-supervised anomaly detection framework, specifically developed for detecting and localizing ECG anomalies, comprises three main components: (i) multi-scale cross-restoration, (ii) trend-assisted restoration, and (iii) the attribute prediction module.

\noindent \textbf{Multi-scale cross-restoration.} 
The architecture of multi-scale cross-restoration is illustrated in Fig.~\ref{fig:datamethod}a. We begin by analyzing a complete global ECG signal $x_{g}\in \mathbb{R}^D$ and segmenting it into local heartbeats. We then randomly select these heartbeats $x_{l}\in\mathbb{R}^d$ to form a global-local ECG pair for further analysis. This pair is masked—globally across scattered areas and locally over a specific continuous region. After masking, the signals are analyzed using separate global and local encoders to extract relevant features. Emulating the approach of professional cardiologists who consider both the overall ECG context and specific heartbeat details, our model utilizes a self-attention mechanism inspired by Vaswani et al.\cite{vaswani2017attention}, which  integrates global and local features into a cohesive cross-attention feature set through concatenation. These integrated features are then dynamically weighted by the self-attention mechanism, allowing the model to focus on areas of the signal that are most relevant.  Following this integration, the model employs two specialized decoders to reconstruct the global signal $\hat{x}_g$ and local signal $\hat{x}_l$ from the extracted features. 
This reconstruction process not only restores the signals but also generates restoration uncertainty maps $\sigma_g$ and $\sigma_l$, which visually represent the challenges encountered during signal restoration. These maps offer valuable insights into the complexities of the restoration task. To enhance our model's performance in signal restoration, we introduce an uncertainty-aware restoration loss\cite{mao2020uncertainty}:
\begin{equation}
\mathcal{L}_{global}=\sum_{k=1}^D \left\{\frac{(x_{g}^{k}-\hat{x}_{g}^{k})^2}{\sigma^{k}_{g}}+\log \sigma^{k}_{g}\right\},
\quad
\mathcal{L}_{local}=\sum_{k=1}^d \left\{\frac{(x_{l}^{k}-\hat{x}_{l}^{k})^2}{\sigma^{k}_l}+\log \sigma^{k}_{l}\right\}.
\end{equation}  
Specifically, each loss function involves normalizing the first term by its respective uncertainty, while the second term is introduced to prevent the prediction of a large uncertainty for all restoration signal points. Here, the superscript $k$ designates the location of the $k$-th element within the signal.  

\noindent \textbf{Trend assisted restoration.} To enhance the model's proficiency in restoring and interpreting the global ECG signals, we generate a continuous time-series trend denoted as $x_t \in \mathbb{R}^D$, which starts by smoothing the global ECG signal with an averaging window during convolution, followed by employing a difference window to identify changes between consecutive points. An additional autoencoder, designed for trend analysis, helps restore the global ECG by focusing on this trend information. The restored signal $\hat{x}_t$ is decoded from the combination of trend feature and global ECG feature, which is then compared with global signal in Euclidean distance for optimization, \textit{i.e.}, $\mathcal{L}_{trend}=\sum_{k=1}^D (x_{g}^{k}-\hat{x}_{t}^{k})^2$. It is worth noting that, the combination of global feature extracted from the global restoration and trend feature from the trend-assisted restoration process contains the key information of the ECG signal, which is essential for the following attribute prediction and classification.

\noindent \textbf{Attribute prediction module.} In addition to the ECG signal, we leverage the rich information from ECG samples, employing a predictive approach to incorporate this data into our anomaly detection framework. Patient-specific information such as age, gender, and other key attributes is extracted from the samples and transformed into normalized vector $Y_{attr}:=[y_1, ... ,y_m]$ as supervisory signals. The feature combination mentioned in the above section is used to predict these attributes, denoted as $\hat{Y}_{attr}:=[\hat{y}_1, ... ,\hat{y}_m]$, via an auxiliary multi-layer perceptron model. By assessing the model's predictions against the actual attributes using mean square error loss, $\mathcal{L}_{pred}(Y_{attr},\hat{Y}_{attr}) = \frac{1}{m}\sum_{i=1}^m (y_i-\hat{y}_i)^2$, we enhance the model's capability to discern the impact of these attributes on the ECG diagnosis. 

These components collectively constitute the anomaly detection framework, which undergoes optimization through the employed loss function. This framework demonstrates efficacy in identifying and pinpointing anomalies within a binary classification scenario, with any ECGs deviating from the norm classified as anomalies. Complementing this anomaly detection framework is a classification network designed to furnish precise ECG diagnostic outcomes for clinical assessment.

\subsubsection{Supervised classification}
\label{sec:claarch}
After completing self-supervised anomaly detection pretraining, our model is designed to provide multi-label classification outcomes based on input ECG signals, recognizing that multiple anomalies may exist within a single ECG. The pretrained anomaly detection model serves as a feature extractor for ECG signals. This pretrained model can be utilized in two ways: first, the extracted ECG features can be used as inputs for classifier training in a two-stage process; second, the anomaly detection model and classifier can be jointly trained in an end-to-end manner.

To enable multi-label classification of ECG signals, we transform the diagnostic findings extracted from ECG samples using the one-hot encoding technique, resulting in a binary label of fixed length $n$. This label, denoted as $\overline{Y} = [\overline{y}_1,...,\overline{y}_{n}]$, signifies the presence or absence of each of the $n$ distinct ECG types within a single ECG signal, thereby providing the ground-truth supervision signal for classification tasks. In our framework, we incorporate an additional neural network classifier to support supervised learning. This classifier takes as input the same combined features utilized in the attribute prediction process and generates predictions for potential cardiac conditions, represented as $P=[p_1, ..., p_n]$. To align these predictions with the ground-truth labels, we employ the Asymmetric Loss function~\cite{benbaruch2020asymmetric}, a widely recognized optimization loss for multi-label classification tasks, as our classification loss $L_{Class}$, which is expressed mathematically as follows:
\begin{equation}
    L_{Class} = \sum_{k=1}^n -\overline{y}_k (1-p_k)^{\gamma_+}\log(p_k) - (1-\overline{y}_k)(p_{km})^{\gamma_-}\log(1-p_{km}).
\end{equation}
In this equation, $\gamma_+$ and $\gamma_-$ serve as positive and negative focusing weight parameters, respectively, providing enhanced control over the contribution of positive and negative elements to the loss function. Since our focus is on emphasizing the positive elements in the label, which represent the specific anomaly types present in the input ECG signal, $\gamma_-$ is set to be larger than $\gamma_+$. 
This adjustment helps the network learn meaningful features from positive labels, despite their relative infrequency. Additionally, $p_{km}=max(p_k-m,0)$ imposes a hard threshold on easily classified negative types, completely disregarding them when their probability is very low. This forces the model to focus on learning from the more challenging-to-classify anomaly types.

\subsection{Training procedure}

\subsubsection{Training strategy}

During the training phase, the comprehensive loss function for anomaly detection framework optimization is formulated as:
\begin{equation}
\mathcal{L}_{AD} = \mathcal{L}_{global} + \alpha \mathcal{L}_{local}+\beta \mathcal{L}_{trend} + \gamma \mathcal{L}_{pred}, 
\end{equation}
where $\alpha$, $\beta$, and $\gamma$ serve as trade-off parameters, weighting the contributions of the individual loss components. For simplicity, we adopt the default values $\alpha = \beta = \gamma = 1.0$. While the loss function for classifier optimization is the asymmetric loss for multi-label classification, denoted as $\mathcal{L}_{Class}$. We explore three distinct training strategies for our framework, including individual training, two-stage training and joint training.

\noindent \textbf{Individual training.} Both the anomaly detection model and the classification model receive the ECG signal as input and are optimized separately by their respective loss functions, $\mathcal{L}_{AD}$ and $\mathcal{L}_{Class}$. There is no overlap in the training process, and the training outcome of the anomaly detection model does not affect the training of the classifier, serving as a baseline model.

\noindent \textbf{Two-stage training.} To explore the impact of the anomaly detection model on the diagnostic outcome, we integrate a pretrained anomaly detection model into the diagnostic framework. Initially, the anomaly detection model processes the ECG signal as input and is optimized by $\mathcal{L}_{AD}$. In the subsequent stage, the anomaly detection model remains fixed while the classifier utilizes the ECG features extracted by the anomaly detection model for optimization, guided by $\mathcal{L}_{Class}$. Although the training processes do not overlap, the training outcome of the anomaly detection model influences the training of the classifier.

\noindent \textbf{Joint training.} We propose a joint training strategy that combines the anomaly detection model and classifier. The ECG signal is initially processed by the anomaly detection model, and features extracted by this model serve as input for the classifier. The joint training objective consists of the combined loss $\mathcal{L}_{AD}+\mathcal{L}_{Class}$, facilitating the simultaneous optimization of both the anomaly detection model and the classifier.

\subsubsection{Training details}

\noindent \textbf{Signal preprocessing.}  
The ECG undergoes pre-processing through a Butterworth filter and Notch filter~\cite{van2019heartpy} to eliminate high-frequency noise and mitigate ECG baseline wander. R-peaks are identified utilizing an adaptive threshold methodology outlined in paper~\cite{van2019rpeak}, which does not necessitate the incorporation of learnable parameters. Subsequently, the positions of the identified R-peaks are employed to segment the ECG sequence, delineating a series of individual heartbeats.

\vspace{3pt} \noindent \textbf{Implementation.}
For the anomaly detection model, we employ a convolutional-based autoencoder architecture, as outlined in paper~\cite{liu2022time}. For the classification model, we utilize the widely employed neural network ResNet, tailored for 1-D data classification tasks. The model undergoes training using the AdamW optimizer, initializing with a learning rate of 1e-4 and a weight decay coefficient of 1e-5, over a span of 50 epochs on a singular NVIDIA GTX 3090 GPU. Decay scheduling is implemented through a single cycle of cosine learning rate. The batch size is configured to 32.

\subsection{Evaluation}

We employ several evaluation metrics to assess the applicability of our method in real-world clinical scenarios. We compute the Area Under the Receiver Operating Characteristic Curve (AUROC) along with Sensitivity and Specificity for each cardiac type. These metrics are then averaged across all cardiac types to provide a comprehensive assessment. For anomaly detection, we further employ the F1 score and precision with a fixed recall rate of 90\%, along with the dice coefficient for localization evaluation. Higher values in these metrics indicate a more effective approach.

To evaluate the classification performance enhanced by anomaly detection model, we investigate three experimental settings: training a classifier only, training a classifier with a fixed anomaly detection model, and joint training of the anomaly detection model and classifier. These three experimental settings are illustrated in Fig.~\ref{fig:datamethod_b}. For anomaly detection and localization, we benchmark our methodology against four state-of-the-art methods: TranAD~\cite{tranad}, AnoTran~\cite{xu2022anomaly}, BeatGAN~\cite{liu2022time} and TSL~\cite{zheng2022task}. 

\textbf{Ethics statement.} The study protocol was approved by the Ruijin Hospital Ethics Committee, Shanghai Jiao Tong University School of Medicine (reference number: 2021-23), and conducted in accordance with the Declaration of Helsinki~\cite{world2013world}.

\subsubsection{Evaluation process}

During the testing phase, for each ECG $x_{test}$, local ECGs are iteratively selected from the segmented heartbeats $\{x_{l,m},m=1,...,M\}$, rather than being randomly chosen as in the training phase. The selected local ECG is combined with the global test ECG to form an input pair for our framework. This input pair is then processed by the framework to generate an anomaly score for detection, a score map for localization, and a classification result for clinical diagnosis.
 
The score map $S(x_{test})$ is derived by combining the predicted outcomes from both the global and local branches, denoted as  $S_g(x_{test})$ and $S_l(x_{test})$, respectively. This combination is mathematically expressed as: $S(x_{test}) = S_g(x_{test}) + S_l(x_{test})$.
In this formulation, the predicted score map $S_g(x_{test})$ for the global branch is defined by:
\begin{equation}
S_g(x_{test}) =  \frac{(x_{test}-\hat{x}_{test})^2}{\sigma_g} + (x_{test}-\hat{x}_{t})^2.
\end{equation}

For the local branch, the predicted score map is obtained by summing the score maps of each heartbeat with their corresponding original positions in the global ECG. This process is mathematically formulated as:
\begin{equation}
S_l(x_{test}) = \sum_{m=1}^M I_m \odot \frac{(x_{l,m}-\hat{x}_{l,m})^2}{\sigma_{l,m}}, 
\end{equation}
where $I_m \in \mathbb{R}^D$ is a zero-one indicator vector, assuming a value of one solely at the position of the $m$-th segmented heartbeat.
The anomaly score $A(x_{test})$  for the entire image is defined as the mean value of all signal points in the score map $S(x_{test})$. 
The classification result $Y(x_{test})$  is a fixed-length vector, with each position corresponding to a specific type of anomaly, representing the probability of the respective particular cardiac anomaly. In the score map, anomaly score, and classification result, higher values indicate a greater likelihood of anomalies.

\subsubsection{Evaluation protocols}

The evaluation metrics employed in our paper include patient/signal point level AUC (area under the Receiver Operating Characteristic curve), F1 score, sensitivity, specificity, precision (@recall=90\%), and dice.
\begin{itemize}
\item AUC is a widely used metric to evaluate the performance of binary classification models. It quantifies the ability of a model to discriminate between positive and negative classes across various threshold values. A higher AUC value indicates better discrimination capability, with a value of 1 representing perfect classification and 0.5 representing random guessing.

\item Sensitivity, also known as true positive rate or recall, measures the proportion of actual positive cases that are correctly identified by the model. It is calculated as
$Sensitivity =\frac{TP}{TP+FN}$. Higher sensitivity indicates that the model effectively captures positive cases.

\item Specificity measures the proportion of actual negative cases that are correctly identified by the model. It is calculated as $Specificity = \frac{TN}{TN+FP}$. Higher specificity indicates that the model effectively excludes negative cases.

\item Precision, also known as positive predictive value, represents the proportion of correctly predicted positive cases out of all instances predicted as positive, $Precision = \frac{TP}{TP+FP}$. Precision calculated at a specific recall level (in this case, at a recall of 90\%) provides insight into the model's performance when a higher recall threshold is desired.

\item The F1 score is a measure of a model's accuracy that considers both precision and recall. It is the harmonic mean of precision and recall, calculated as $F1 = \frac{2\times Precision \times Recall}{Precision + Recall}$. F1 score ranges from 0 to 1, with higher values indicating better model performance in terms of both precision and recall.

\item The Dice coefficient is a measure of similarity between two sets, quantifying the overlap between the predicted and ground truth anomaly masks. It is calculated as $Dice = 2\times \frac{Intersection}{Union}$, where the intersection represents the overlapping area between the predicted and ground truth masks, and the union represents the total area encompassed by both masks.
\end{itemize}
For these evaluation protocols, $TP$, $TN$, $FP$, $FN$ represent true positives, true negatives, false positives, and false negatives, respectively.

\subsubsection{Prospective evaluation}
To assess the practical efficacy of our method, we conducted prospective validation using ECG data from the emergency department, wherein clinicians were tasked with making rapid diagnoses under high-pressure conditions, typically within a timeframe of less than one minute. The utilized data comprised 238 distinct ECGs, representing 50 unique ECG types. After being diagnosed by three doctors under different settings, an adjudicating clinician with the highest level of expertise reviewed and confirmed the diagnoses provided by these clinicians as the final conclusions. Using this final conclusion as the gold standard, we established four comparison groups: clinicians' diagnosis results under time limitation, clinicians' diagnosis results without time limitation, our model's diagnosis results, and clinicians' diagnosis results with our model's assistance.

The evaluation of results was conducted along three dimensions: accuracy, efficiency, and completeness. Accuracy, being the most critical metric, was assessed by comparing the diagnosis results across four comparison groups with the final conclusion. Specifically, if all ECG types for an ECG sample in the final conclusion were accurately recorded, it was deemed a correctly diagnosed sample; otherwise, it was classified as a misdiagnosed sample. Subsequently, accuracy metrics were computed for the four distinct comparison groups. Efficiency was evaluated by the average diagnosis time for each ECG, with our model running on a single NVIDIA GTX 3090 GPU. Finally, the completeness of the diagnosis conclusion was evaluated based on whether the conclusion included detailed signal information in addition to key diseases.

\subsection{Related work} 
In the realm of ECG analysis, current research progresses along two primary avenues: computer-aided diagnosis and anomaly detection. The former, driven by advancements in deep learning, has emerged as the predominant approach. Conversely, the latter avenue, although less explored, presents novel opportunities for addressing long-tail challenges in diagnosis. Our study represents the inaugural effort to integrate these two approaches, aiming to enhance ECG diagnosis particularly in cases with long-tail distribution of anomalies.

\begin{itemize}
\setlength\itemsep{0.15cm}
\item \textbf{Computer-aided ECG diagnosis.}
The advancement of deep learning has greatly enhanced computer-aided ECG diagnosis performance~\cite{liu2021deep}, transitioning from the conventional rule-based diagnostic approach~\cite{EBRAHIMI2020100033}. However, current researches following classification framework require a large amount of labeled abnormal data~\cite{wagner2020ptb} and are limited to detecting only the specific anomaly types provided in the training data, such as arrhythmia classification~\cite{wang2023arrhythmia, rahul2022automatic}, atrial fibrillation classification~\cite{nankani2022atrial, feng2022novel}, or multi-label classification~\cite{du2021fm, cao2021practical}, thereby falling short of identifying all unseen abnormal ECG signals. This constraint diminishes their practical utility in clinical diagnosis.

\item \textbf{Anomaly detection in ECG.}
The scarcity and diversity of anomaly types, combined with the prevalence of normal ECGs, pose challenges due to the limited availability of labeled anomalies for supervised training~\cite{ukil2016iot}. Consequently, anomaly detection tasks are typically approached as unsupervised learning tasks, relying on utilizing normal samples for training and identifying any samples that deviate from this norm as anomalies~\cite{pang2021deep}. The objective of object anomaly detection is to effectively distinguish between normal and anomalous samples, which can be viewed as a binary classification problem~\cite{chandola2009anomaly}. Current research frames ECG anomaly detection within the broader scope of time-series anomaly detection, primarily focusing on two key approaches: reconstruction-based methods and self-supervised~\cite{krishnan2022self} learning-based methods. Reconstruction-based methods~\cite{xu2022anomaly, tranad, zhou2019beatgan} employ generative neural networks~\cite{goodfellow2020generative} to reconstruct normal samples, operating under the assumption that a model trained on normal samples will struggle to accurately reconstruct abnormal regions. While self-supervised learning-based methods~\cite{zheng2022task} explore the use of proxy tasks to improve the representation learning of normal samples.
\end{itemize}

\section{Data availability}
ECG-LT data is not publicly accessible as it is internal data of patients of Ruijin Hospital, Shanghai Jiao Tong University School of Medicine in Shanghai, China. The study protocol was approved by the Ruijin Hospital Ethics Committee, Shanghai Jiao Tong University School of Medicine (reference number: 2021-23), and conducted in accordanc with the Declaration of Helsinki

PTB-XL data is available at \href{https://physionet.org/content/ptb-xl/1.0.3/}{https://physionet.org/content/ptb-xl/1.0.3/}.

\section{Code availability}
Source code, data samples, and pre-trained models of this
work will be uploaded to \href{https://github.com/MediaBrain-SJTU/ECGAD}{https://github.com/MediaBrain-SJTU/ECGAD}.

\clearpage

\bibliographystyle{sn-mathphys} 
\bibliography{references} 

\captionsetup[figure]{name=Extended Figure }
\captionsetup[table]{name=Extended Table }
\setcounter{figure}{0}
\setcounter{table}{0}

\clearpage

\section{Extended Data Figures and Tables}

\begin{table}[!htb]
\caption{Detailed information about common data on our general hospital ECG-LT dataset.}
\centering
\scalebox{0.9}{
\begin{tabular}{C{1.7cm}|C{1.4cm}|C{8.0cm}C{2.2cm}C{2.2cm}} 
\hline
Data  & Condition & \multirow{2}{*}{Cardiac Conditions} & \multirow{2}{*}{\# All data} & \multirow{2}{*}{\# Test data}\\
Type & Types & & & \\
\hline
\multirow{37}{*}{Common}  & \multirow{37}{*}{37} & Normal electrocardiogram & 539,607 & 94,304\\
& & T wave changes & 182,756 & 33,907\\
& & Sinus bradycardia & 99,275 & 17,463\\
& & ST-T segment changes & 80,334 & 11,953\\
& & Sinus tachycardia & 61,287 & 11,272\\
& & ST segment changes & 49,129 & 7,177\\
& & Mild T wave changes & 45,904 & 6,692\\
& & First-degree atrioventricular block & 35,470 & 6,774\\
& & Left ventricular high voltage & 35,104 & 5,750\\
& & Atrial fibrillation & 35,090 & 95\\
& & Complete right bundle branch block & 34,614 & 6,161\\
& & Atrial premature beat & 29,376 & 5,684\\
& & Sinus arrhythmia & 27,141 & 4,277\\
& & Ventricular premature beat & 25,765 & 4,401\\
& & Low voltage & 21,759 & 4,103\\
& & Left anterior fascicular block & 20,035 & 3,842\\
& & Mild ST segment changes & 17,542 & 3,168\\
& & ST segment saddleback elevation & 16,898 & 2,698\\
& & Incomplete right bundle branch block & 15,917 & 3,045\\
& & ST segment depression & 13,246 & 1,101\\
& & Peaked T wave & 9,979 & 2,538\\
& & Clockwise rotation & 6,786 & 1,264\\
& & Atrial flutter & 6,248 & 39\\
& & Prominent U wave & 5,788 & 1,496\\
& & Complete left bundle branch block & 5,419 & 812\\
& & Ventricular paced rhythm & 5,055 & 474\\
& & Atrial tachycardia & 5,014 & 612\\
& & Intraventricular conduction block & 4,830 & 738\\
& & Left ventricular hypertrophy & 4,423 & 731\\
& & Paroxysmal atrial tachycardia & 4,185 & 598\\
& & Inferior myocardial infarction & 3,346 & 423\\
& & Mild ST-T segment changes & 3,174 & 405\\
& & Frequent atrial premature beat & 3,008 & 424\\
& & Extensive anterior myocardial infarction & 2,988 & 485\\
& & Anteroseptal Myocardial Infarction & 2,723 & 481\\
& & Frequent ventricular premature beat & 2,604 & 360\\
& & Abnormal Q wave & 2,539 & 223\\
\hline
\end{tabular}
}
\label{tab:dataset_common}
\end{table}

\begin{table}[!p]
\caption{Detailed information about uncommon data on our general hospital ECG-LT dataset.}
\centering
\scalebox{0.9}{
\begin{tabular}{C{1.7cm}|C{1.4cm}|C{8.0cm}C{2.2cm}C{2.2cm}} 
\hline
Data  & Condition & \multirow{2}{*}{Cardiac Conditions} & \multirow{2}{*}{\# All data} & \multirow{2}{*}{\# Test data}\\
Type & Types & & & \\
\hline
\multirow{36}{*}{Uncommon}  & \multirow{36}{*}{36} & Prolonged QT interval & 2,486 & 410\\
& & Right axis deviation & 2,476 & 450\\
& & Old anterior myocardial infarction & 2,403 & 418\\
& & Old inferior myocardial infarction & 2,344 & 362\\
& & Left axis deviation & 1,856 & 188\\
& & Atrial paced rhythm & 1,854 & 112\\
& & Pre-excitation syndrome & 1,829 & 292\\
& & Intraventricular conduction delay & 1,740 & 269\\
& & Flat T wave & 1,701 & 128\\
& & Poor R wave progression or reversed progression & 1,642 & 281\\
& & Non-conducted atrial premature beat & 1,445 & 256\\
& & Junctional escape beat & 1,444 & 142\\
& & Insertional ventricular premature beat & 1,351 & 268\\
& & Short PR interval with normal QRS complex & 1,299 & 183\\
& & Second-degree atrioventricular block & 1,287 & 217\\
& & Couplet atrial premature beat & 1,259 & 206\\
& & Atrial bigeminy & 1,244 & 223\\
& & Paroxysmal supraventricular tachycardia & 1,107 & 5\\
& & Elevated J point & 985 & 162\\
& & Paced electrocardiogram & 972 & 1\\
& & Peaked P wave & 897 & 196\\
& & Couplet ventricular premature beat & 744 & 114\\
& & Non-paroxysmal junctional tachycardia & 743 & 81\\
& & Bifid P wave & 732 & 193\\
& & Third-degree atrioventricular block & 698 & 15\\
& & Second-degree type 1 atrioventricular block & 692 & 137\\
& & Horizontal ST segment depression & 679 & 48\\
& & Junctional premature beat & 620 & 115\\
& & Ventricular tachycardia & 542 & 52\\
& & Ventricular escape beat & 509 & 30\\
& & Anterior myocardial infarction & 498 & 38\\
& & Inverted T wave & 484 & 51\\
& & Posterior myocardial infarction & 459 & 64\\
& & Long RR interval & 441 & 3\\
& & Atrial trigeminy & 436 & 77\\
& & Paroxysmal ventricular tachycardia & 428 & 51\\
\hline
\end{tabular}
}
\label{tab:dataset_uncommon}
\end{table}

\begin{table}[p]
\caption{Detailed information about rare data on our general hospital ECG-LT dataset.}
\centering
\scalebox{0.9}{
\begin{tabular}{C{1.7cm}|C{1.4cm}|C{8.0cm}C{2.2cm}C{2.2cm}}  
\hline
Data  & Condition & \multirow{2}{*}{Cardiac Conditions} & \multirow{2}{*}{\# All data} & \multirow{2}{*}{\# Test data}\\
Type & Types & & & \\
\hline
\multirow{43}{*}{Rare}   & \multirow{43}{*}{43} & High lateral myocardial infarction & 394 & 30\\
& & Upsloping ST segment depression & 368 & 30\\
& & Downsloping ST segment depression & 311 & 33\\
& & Second-degree sinoatrial block & 257 & 50\\
& & High-degree atrioventricular block & 247 & 20\\
& & Right ventricular myocardial infarction & 213 & 10\\
& & Dextrocardia & 167 & 23\\
& & Right ventricular hypertrophy & 164 & 18\\
& & Ventricular bigeminy & 160 & 22\\
& & Second-degree type 2 sinoatrial block & 117 & 21\\
& & Second-degree type 1 sinoatrial block & 117 & 27\\
& & Atrial tachycardia with variable conduction & 115 & 0\\
& & Sinus arrest & 109 & 12\\
& & Biphasic P wave & 106 & 17\\
& & Lateral myocardial infarction & 101 & 12\\
& & Old lateral myocardial infarction & 86 & 7\\
& & Second-degree type 2 atrioventricular block & 85 & 15\\
& & Left posterior fascicular block & 80 & 4\\
& & Biphasic T wave & 65 & 9\\
& & Muscle artifact & 61 & 15\\
& & Old posterior myocardial infarction & 61 & 10\\
& & Inverted P wave & 55 & 7\\
& & Lead detachment & 51 & 0\\
& & Sinus node wandering rhythm & 48 & 5\\
& & Unstable baseline & 47 & 5\\
& & Ventricular trigeminy & 46 & 5\\
& & Subendocardial myocardial infarction & 40 & 3\\
& & Old high lateral myocardial infarction & 37 & 1\\
& & Atrial escape beat & 33 & 5\\
& & Atrial arrhythmia & 30 & 0\\
& & Broadened P wave & 23 & 1\\
& & Insertional atrial premature beat & 20 & 6\\
& & Interference atrioventricular dissociation & 20 & 3\\
& & Counterclockwise rotation & 19 & 4\\
& & Left atrial hypertrophy & 19 & 1\\
& & Right atrial hypertrophy & 19 & 0\\
& & Paroxysmal junctional tachycardia & 17 & 1\\
& & Ventricular quadrigeminy & 13 & 2\\
& & Ventricular fibrillation & 7 & 0\\
& & Shortened QT interval & 6 & 6\\
& & Alternating left and right bundle branch block & 6 & 1\\
& & Baseline drift & 4 & 1\\
& & Biauricular hypertrophy & 1 & 0\\
\hline
\end{tabular}
}
\label{tab:dataset_rare}
\end{table}

\begin{figure}[p]
\centering
\includegraphics[width=1.0\textwidth]{figure/data_type.pdf}
\caption{\textbf{Comprehensive analysis of the novel ECG-LT dataset.} 
\textbf{a.} Hierarchical architecture of cardiac types.
\textbf{b.} Comparison of the number of cardiac types in the ECG-LT dataset to those in existing ECG databases.
\textbf{c.} Age distribution across the training, internal validation, and external validation sets. 
\textbf{d.} Gender distribution across the training, internal validation, and external validation sets.
}
\label{fig:datainfo}
\end{figure}

\begin{figure}[p]
\centering
\includegraphics[width=1.0\textwidth]{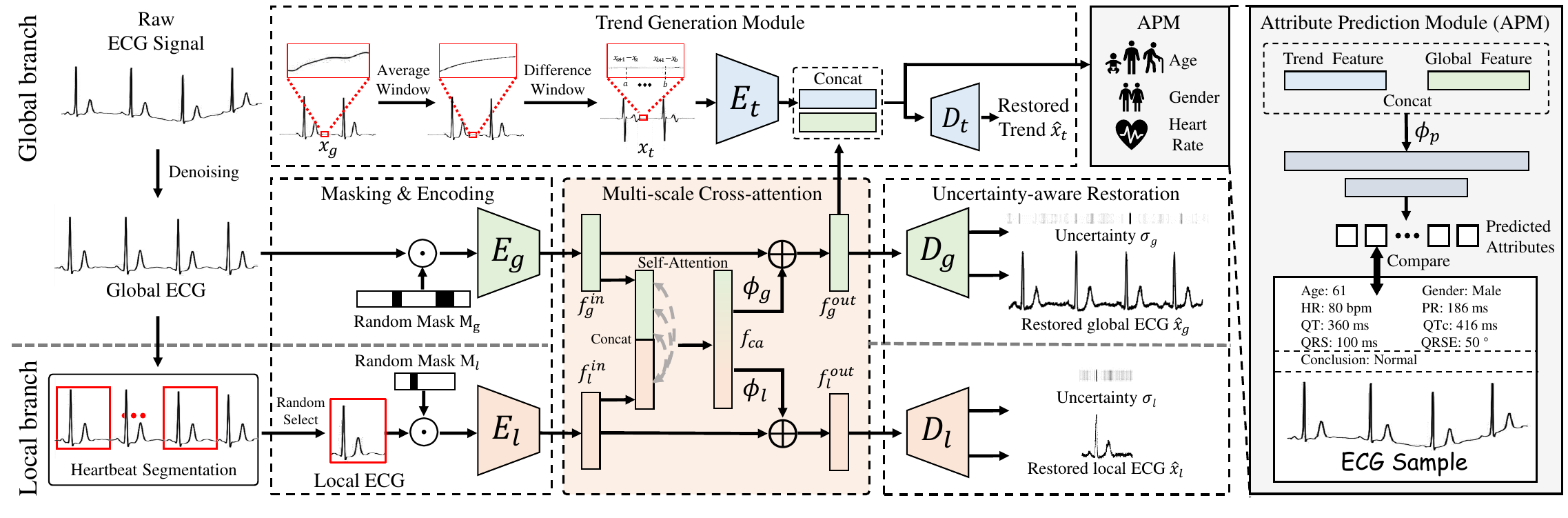}
\caption{The details of multi-scale cross-restoration framework for ECG anomaly detection.}
\label{fig:model_supp}
\end{figure}

\begin{figure}[!htb]
    \centering
    \includegraphics[width = \textwidth]{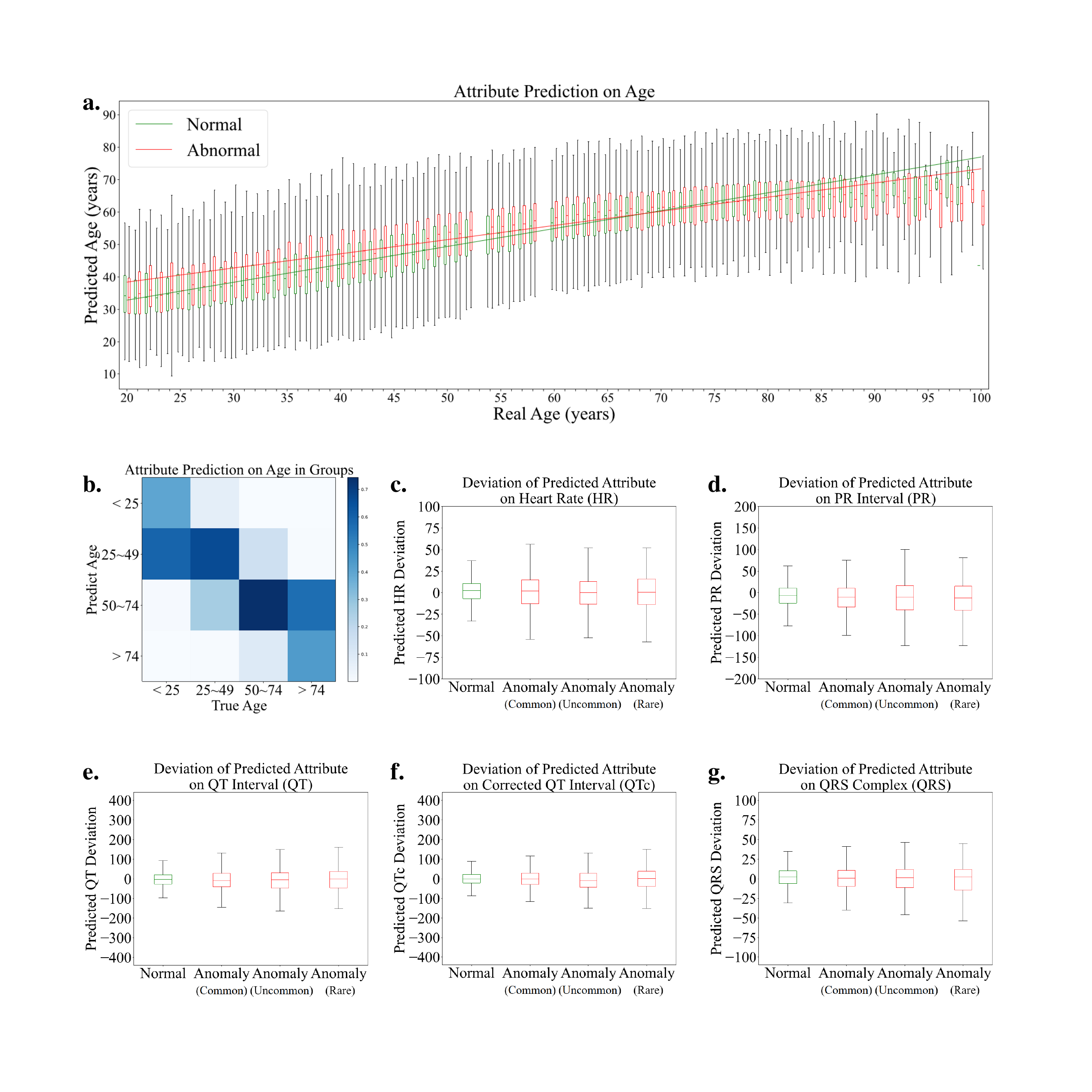}
    \caption{\textbf{Evaluation of the attribute prediction module.} \textbf{a.} Comparison of predicted ages across different age groups for normal and abnormal ECGs. \textbf{b.} Age classification accuracy for normal ECGs depicted across age ranges, showing predicted versus actual age. \textbf{c.} Heart rate prediction deviation compared to a standard reference range (y-axis) across varying anomaly rarity (x-axis). \textbf{d-g.} Analysis of deviations in PR interval, QT interval, corrected QT interval, and QRS complex predictions.}
\label{fig:abl_apm}
\end{figure}

\begin{table}[!htb]
\caption{Detailed information about data in the prospective validation.}
\centering
\scalebox{0.9}{
\begin{tabular}{C{1.7cm}|C{1.5cm}|C{8.0cm}C{2.5cm}C{2.5cm}} 
\hline
Data  & Condition & \multirow{2}{*}{Cardiac Conditions} & \multirow{2}{*}{\#Data} & \multirow{2}{*}{Rarity}\\
Source & Types & &  & \\
\hline
   & \multirow{50}{*}{50} & Normal electrocardiogram & 40 & Common\\
& & Sinus tachycardia & 35 & Common\\
& & Atrial fibrillation & 26 & Common\\
& & ST-T segment changes & 26 & Common\\
& & T wave changes & 24 & Common\\
& & Sinus bradycardia & 21 & Common\\
& & Complete right bundle branch block & 18 & Common\\
& & Low voltage & 18 & Common\\
& & Left ventricular high voltage & 15 & Common\\
& & Atrial flutter & 15 & Common\\
& & Atrial premature beat & 12 & Common\\
& & ST segment changes & 12 & Common\\
& & Ventricular premature beat & 11 & Common\\
& & First-degree atrioventricular block & 11 & Common\\
& & Left anterior fascicular block & 10 & Common\\
& & Paroxysmal supraventricular tachycardia & 9 & Uncommon\\
& & Peaked T wave & 9 & Common\\
& & Sinus arrhythmia & 9 & Common\\
& & Prominent U wave & 9 & Common\\
& & Intraventricular conduction block & 9 & Common\\
& & Mild T wave changes & 9 & Common\\
& & Ventricular paced rhythm & 7 & Common\\
& & ST segment saddleback elevation & 7 & Common\\
& & Incomplete right bundle branch block & 6 & Common\\
Emergency & & Mild ST segment changes & 6 & Common\\
department & & Extensive anterior myocardial infarction & 5 & Common\\
& & Complete left bundle branch block & 5 & Common\\
& & Atrial tachycardia & 5 & Common\\
& & Ventricular escape beat & 5 & Uncommon\\
& & Old anterior myocardial infarction & 4 & Uncommon\\
& & Intraventricular conduction delay & 4 & Uncommon\\
& & Poor R wave progression or reversed progression & 4 & Uncommon\\
& & Frequent ventricular premature beat & 4 & Common\\
& & Frequent atrial premature beat & 4 & Common\\
& & Prolonged QT interval & 4 & Uncommon\\
& & Paced electrocardiogram & 4 & Uncommon\\
& & Couplet atrial premature beat & 3 & Uncommon\\
& & Left ventricular hypertrophy & 3 & Common\\
& & Old inferior myocardial infarction & 3 & Uncommon\\
& & Paroxysmal atrial tachycardia & 3 & Common\\
& & Junctional escape beat & 3 & Uncommon\\
& & Short PR interval with normal QRS complexs & 2 & Uncommon\\
& & ST segment depression & 2 & Common\\
& & Clockwise rotation & 2 & Common\\
& & Junctional premature beat & 1 & Uncommon\\
& & Non-conducted atrial premature beat & 1 & Uncommon\\
& & Elevated J point & 1 & Uncommon\\
& & Flat T wave & 1 & Uncommon\\
& & Left axis deviation & 1 & Uncommon\\
& & Pre-excitation syndrome & 1 & Uncommon\\
\hline
\end{tabular}
}
\label{tab:dataset_cohort}
\end{table}

\begin{figure}[t]
\centering
\includegraphics[width=0.8\textwidth]{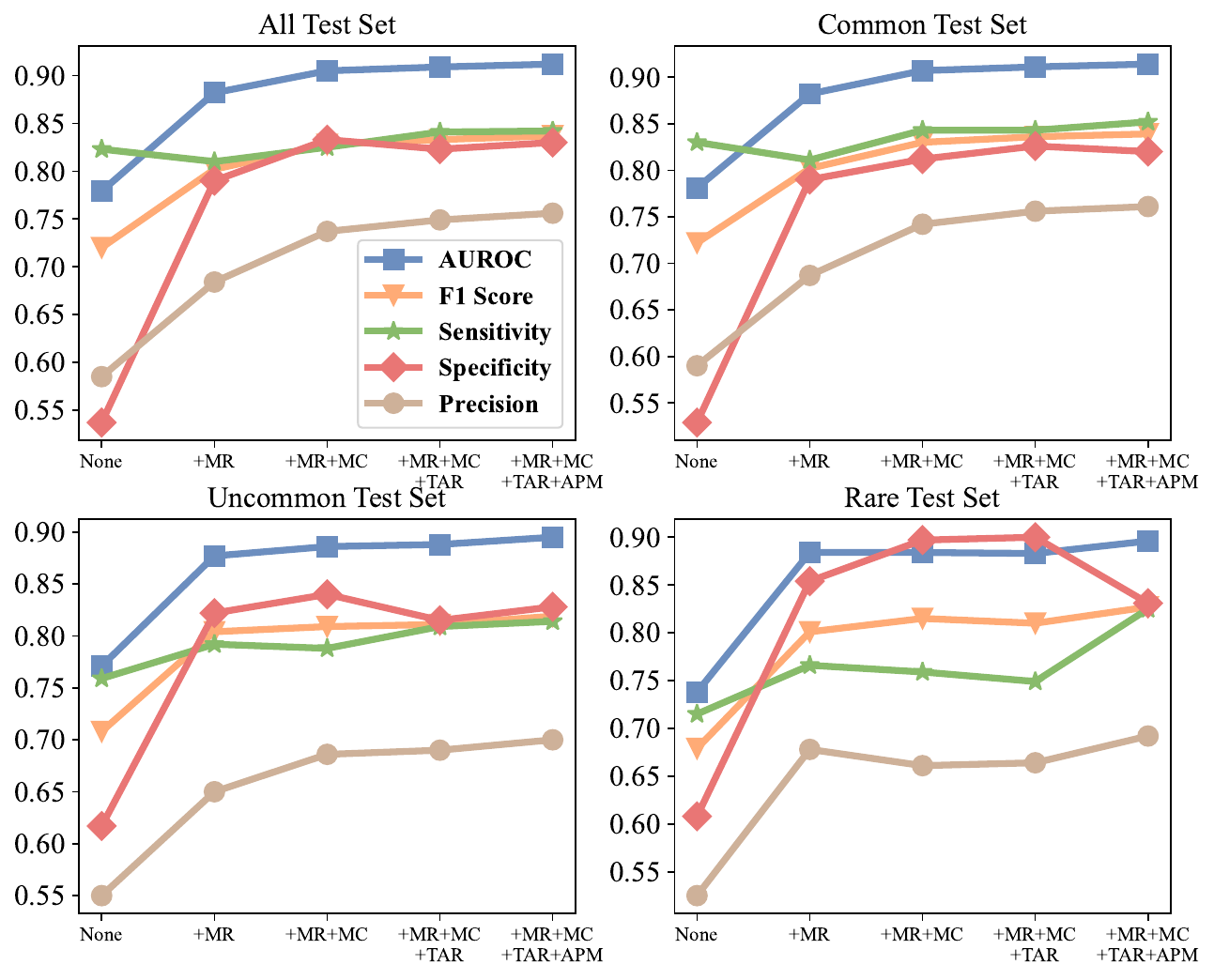}
\caption{Ablation studies evaluating the impact of masking and restoring (MR), multi-scale cross-attention (MC), trend assisted restoration (TAR), and the attribute prediction module (APM) on various test sets.}
\label{fig:abl_ruijin}
\end{figure}

\begin{table}[t]
 \caption{Attribute prediction results for various attributes and data types on general hospital ECG dataset. Results are shown in accuracy for binary gender and averaged deviation for other attributes.}
\centering
\scalebox{0.9}{
\setlength{\tabcolsep}{1.3pt}{
\begin{tabular}{C{4.5cm}|C{3.5cm}|C{1.8cm}C{1.8cm}C{1.8cm}C{1.8cm}C{1.8cm}} 
\hline
\multirow{2}{*}{Attribute} & \multirow{2}{*}{Reference Range} & \multirow{2}{*}{Normal} & \multicolumn{4}{c}{Abnormal} \\
& & & All & Common & Uncommon & Rare \\
\hline
Gender & 0 (male) or 1 (female)  & 86.5\% Acc               & 75.9\% Acc           &  76.2\% Acc         & 69.2\% Acc   & 70.9\% Acc \\
Age & 0 $\sim$ 100 years & $\pm$12.6 & $\pm$14.0 & $\pm$13.8 & $\pm$16.8 & $\pm$18.1\\
Heart Rate (HR) & 60 $\sim$ 100 bpm & $\pm$3.64 & $\pm$6.94 & $\pm$6.86 & $\pm$8.20 & $\pm$8.62\\
PR Interval (PR) & 120 $\sim$ 200 ms & $\pm$18.5 & $\pm$26.7 & $\pm$26.2 & $\pm$35.2 & $\pm$37.8\\
QT Interval (QT) & 320 $\sim$ 440 ms & $\pm$20.7 & $\pm$33.0 & $\pm$32.4 & $\pm$43.2 & $\pm$44.0\\
Corrected QT Interval (QTc) & 350 $\sim$ 440 ms & $\pm$21.2 & $\pm$32.7 & $\pm$32.1 & $\pm$42.3 & $\pm$46.4\\
QRS Complex (QRS) & 60 $\sim$ 110 ms & $\pm$5.74 & $\pm$9.76 & $\pm$9.62 & $\pm$12.3 & $\pm$13.6\\
\hline
\end{tabular}
}}
\label{tab:abl_apvalue}
\end{table}

\end{document}